\documentclass{article}

\usepackage{PrimeArxiv}
\usepackage[numbers]{natbib}
\usepackage{amsmath}

\usepackage[utf8]{inputenc} 
\usepackage[T1]{fontenc}    
\usepackage{amssymb}
\usepackage{upgreek}
\DeclareMathAlphabet{\mathcal}{OMS}{cmsy}{m}{n} 
\usepackage{fancyhdr}       
\usepackage{graphicx}       
\usepackage[capitalise]{cleveref}
\crefname{figure}{Figure}{Figures}
\Crefname{figure}{Figure}{Figures}
\usepackage{url}            
\usepackage{listings}
\usepackage{setspace}

\title{Grammar-based Ordinary Differential Equation Discovery
}

\author{
  Karin L. Yu \\
  ETH AI Center and \\
  Institute of Structural Engineering \\
  ETH Zurich \\
  Switzerland\\
  \texttt{karin.yu@ibk.baug.ethz.ch}\\
  \And
  Eleni Chatzi \\
  Institute of Structural Engineering \\
  ETH Zurich \\
  Switzerland\\
  \texttt{chatzi@ibk.baug.ethz.ch} \\
  \And
  Georgios Kissas \\
  ETH AI Center \\
  ETH Zurich \\
  Switzerland\\
  \texttt{gkissas@ai.ethz.ch} \\
}

\begin{document}
\maketitle

\begin{abstract}
The understanding and modeling of complex physical phenomena through dynamical systems has historically driven scientific progress, as it provides the tools for predicting the behavior of different systems under diverse conditions through time. The discovery of dynamical systems has been indispensable in engineering, as it allows for the analysis and prediction of complex behaviors for computational modeling, diagnostics, prognostics, and control of engineered systems. Joining recent efforts that harness the power of symbolic regression in this domain, we propose a novel framework for the end-to-end discovery of ordinary differential equations (ODEs), termed Grammar-based ODE Discovery Engine (GODE). The proposed methodology combines formal grammars with dimensionality reduction and stochastic search for efficiently navigating high-dimensional combinatorial spaces. Grammars allow us to seed domain knowledge and structure for both constraining, as well as, exploring the space of candidate expressions. GODE proves to be more sample- and parameter-efficient than state-of-the-art transformer-based models and to discover more accurate and parsimonious ODE expressions than both genetic programming- and other grammar-based methods for more complex inference tasks, such as the discovery of structural dynamics. Thus, we introduce a tool that could play a catalytic role in dynamics discovery tasks, including modeling, system identification, and monitoring tasks.
\end{abstract}


\section{Introduction}
\label{ch1:intro}

Scientific discovery unfolds as an iterative cycle \cite{kuhn_structure_1997} in which mathematical formalizations, such as differential equations, are derived from fundamental principles to capture and explain experimental observations. Not only do these mathematical models provide a precise language for expressing complex physical phenomena, but also predict the behavior of systems under various conditions and constraints \cite{kissas_language_2024}. By solving these models under different initial conditions, researchers can forecast how systems behave in space and time, and these predictions can then be rigorously tested against experimental data \cite{schmidt_distilling_2009}. When the experimental outcomes confirm the predictions, the model is validated and becomes a powerful tool for understanding nature; conversely, discrepancies between theory and experiment drive further refinements of the models or even inspire the development of entirely new theoretical frameworks \cite{popper_logic_1959}. This dynamic interplay between mathematical theory and empirical validation has been pivotal in advancing our understanding of the natural world, continually shaping progress in science and engineering while encouraging an ever-deepening exploration of the underlying laws of nature \cite{morgan_models_1999}. 

Inferring mathematical formalizations of processes is essential for robust modeling, estimation, monitoring, and control of dynamical systems in a range of scientific fields, including physics, chemistry, biology, and engineering \cite{wang_scientific_2023}. However, extracting differential equations from first principles, especially for nonlinear systems, exhibits complex behaviors that often defy simple analytical approximations, making it difficult to derive their governing equations \cite{bongard_automated_2007}. To address these issues, symbolic regression, a powerful method for uncovering mathematical relationships in data, approaches have been proposed to enable the extraction of differential equations directly from sparse observations \cite{brunton_discovering_2016, dascoli_odeformer_2023, omejc_probabilistic_2024}. 

Symbolic regression, a cornerstone of data-driven scientific discovery, can be understood through a taxonomy that organizes methods into two broad categories, each reflecting a distinct approach for uncovering the underlying mathematical relationships in data. \emph{Discrete iterative algorithms} initiate with a set of building blocks or primitives, such as unary and binary operators or subexpressions. They then systematically explore the combinatorial space of possible expressions, piecing together these operations in diverse ways until the symbolic expression that best fits the observed data is identified. Different approaches are identified through the possible allowed combinations; for example, \emph{genetic programming} methods \cite{schmidt_distilling_2009, bongard_automated_2007, koza_genetic_1992, tsoulos_solving_2006} allow any combination, syntactically valid or invalid, between primitives. This may lead to an explosion of the combinatorial complexity, resulting in even modern approaches of genetic programming, see PySR as a prime example \cite{cranmer_discovering_2020,cranmer_interpretable_2023}, to fail in associated inference (discovery) tasks. Recent approaches attempt to improve the sampling of the initial populations with neural-guided search \cite{mundhenk_symbolic_2021}, whereas deep symbolic regression uses recurrent neural networks to search in discrete spaces \cite{petersen_deep_2019}. As an alternative to genetic programming and to reduce complexity, \emph{formal grammar}-based approaches \cite{todorovski_declarative_1997}, such as ProGED \cite{omejc_probabilistic_2024, brence_probabilistic_2021}, have been proposed that only allow combinations that result in syntactically valid mathematical expressions, thus reducing the space of possible candidate expressions. An extreme approach to achieve this reduction is \emph{sparse regression} methods, also known as dictionary-based methods. Sparse regression methods assume a pre-established library of subexpressions that capture common functional patterns. The model is restricted to form linear combinations of these subexpressions, and the task is reduced to identifying the optimal weighted sum \cite{brunton_discovering_2016, messenger_weak_2021} that best fits the observed data. Sparse regression methods require a careful selection of candidate subexpressions, which ultimately relies on the expertise of the user. Discrete iterative algorithms are task-agnostic, flexible, and easy to adapt to new scenarios; however, due to the fact that they search a large discrete space for the optimal model, depending on the generality of the algorithm, they can be computationally intensive and slow. 

The other category of symbolic regression methods that we identify is the one we refer to as \emph{direct algorithms}. Direct algorithms learn an operator between the numerical evaluation of a system, for example, the trajectory of an ordinary differential equation (ODE), and its symbolic representation. State-of-the-art direct approaches, such as ODEFormer \cite{dascoli_odeformer_2023} and variants \cite{lample_deep_2019,kamienny_end--end_2022,kamienny_deep_2023,becker_predicting_2023}, are constructed in the following steps. During training, the algorithm samples a 3-tuple of a symbolic expression, a domain, and initial conditions, computes the system trajectory, and learns an operator between a system trajectory and its symbolic representation. At inference time, the model directly provides an initial guess for the expression that best fits the data, and a second optional step is also considered where the scalars of the expression are optimized for a constant skeleton expression. Direct methods are pre-trained on a large corpus of predefined tasks \cite{dascoli_odeformer_2023, lample_deep_2019, dascoli_deep_2022}, which allows for embedding expressions into vectorial representations, known as tokenization \cite{sennrich_neural_2015}, that can be efficiently recovered during discovery. We argue that direct methods inherit specific drawbacks by construction, such as sample and parameter inefficiency, and inflexibility in adapting to new tasks.

We propose a fundamentally different approach in the discovery of ODEs that builds upon ideas from both iterative and direct methods, called Grammar-based ODE Discovery Engine (GODE). More specifically, the proposed methodology considers formal grammars to restrict the space of candidate operators \cite{hopcroft_introduction_1999}, a pre-training strategy, as in direct methods, to embed the candidate expressions into a low-dimensional continuous latent space via the Grammar Variational Autoencoder (GVAE) \cite{kusner_grammar_2017}, and finally stochastic iterative search in the continuous low-dimensional space to discover the ODE that the given measurements or observations satisfy. GODE, compared to direct methods, is more efficient and adaptable to new scenarios; it requires fewer resources as the pre-training only considers symbolic information, the model is comprised of far fewer parameters, and it needs many fewer samples. We show that the proposed methodology, compared to search methods, can discover more complex expressions often encountered in engineering tasks. Moreover, GODE allows for tackling a larger set of symbolic discovery problems, such as implicit ODEs, that it was not previously possible. 

The contributions of this paper can be summarized as follows:
\begin{itemize}

\item[(i)] \textit{Generality:} We propose the grammar-based approach GODE that does not require a targeted design of the candidate model space, making it applicable to a wide variety of cases. Unlike direct methods, which rely on large tailored datasets to capture all specific problem instances, our method uses formal grammars to define ODEs generically, ensuring syntactic validity while supporting broad applicability. Moreover, grammars allow us to introduce structure and domain biases to guide the discovery process.

\item[(ii)] \textit{Task-agnostic:} GODE is task-agnostic as it is first pre-trained on different candidate ODEs and then the symbolic task is defined. This means that the proposed methodology generalizes for different domains and initial conditions.

\item[(iii)] \textit{Sample efficiency:} The proposed embedding methodology is constructed using grammars instead of tokenization which is shown to require far fewer training samples to provide better accuracy than state-of-the-art methods, such as ODEFormer. 

\item[(iv)] \textit{Practicality:} Our end-to-end model reformulates the discovery into a search problem, eliminating the need to solve ODEs during the generation of the training dataset. Furthermore, GODE is not restricted to explicit ODEs, broadening its applicability and adaptability such as incorporating partial information (e.g., forcing terms).

\end{itemize} 
\section{Methodology}
\label{ch2:method}

\begin{figure}[h!]
    \centering
    \includegraphics[scale=0.85]{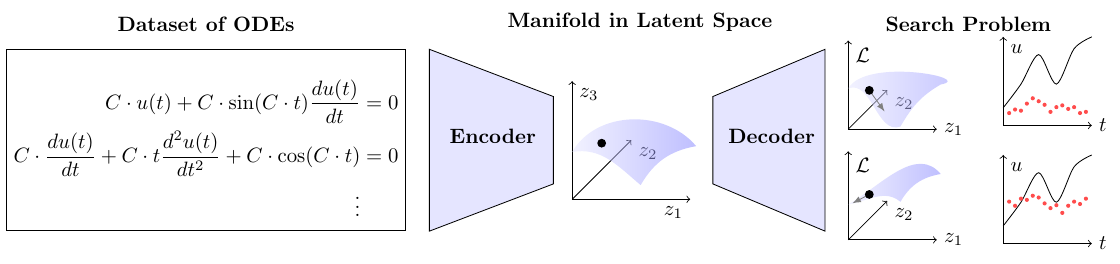}
    \caption{Overview of the end-to-end GODE process with formal grammars: First, grammars allow for an efficient generation of a dataset, then a VAE is trained to embed this dataset into a continuous latent space, which is searched during inference with a stochastic algorithm to identify the best-fitting ODE.}
    \label{fig:ch2:overview}
\end{figure}

We propose the novel approach GODE that combines elements of both \emph{iterative} and \emph{direct} algorithms. GODE is realized in the following steps: A discrete set of candidate ODEs is constructed using grammars, then it is embedded in a continuous low-dimensional manifold using a deep learning model, and lastly, a stochastic search algorithm is defined over the manifold to discover the ODE that the data best satisfy (see \cref{fig:ch2:overview}). We designate the notation $\Box(t)$ to represent the continuous ground truth solution, $\Box_t$ for the discrete ground truth trajectories, $\tilde{\Box}_t$ for the noisy observations, $\Box^{NN}_t$ for the approximated discrete trajectories, and $\hat{\Box}_t$ for the discrete trajectories of the predicted equation.

\subsection{Grammar-generated candidate models}
\label{ch2:subsec:library}

The approach of defining a set of candidate expressions and a structured way to sample from it was first introduced, to the best of our knowledge, by Lample et. al. \cite{lample_deep_2019}, where the authors randomly generated graph trees and assigned labels to the graph nodes to create a combinatorially large number of expressions. In this case, the set of candidate expressions $\mathrm{S}$ is rigorously defined considering $\mathcal{G}= \{ \mathcal{V}, \mathcal{E}, l \}$ a general graph topology, where $\mathcal{V}$ is a set of nodes, $\mathcal{E} \subseteq \mathcal{V} \times \mathcal{V}$ a set of edges, and $ l: \mathcal{V} \rightarrow L$ a labeling function that assigns a label from $L = \{ U, B, C, A \}$ to a node $\mathit{v} \in \mathcal{V}$ \cite{oikonomou_neuro-symbolic_2025}. The subset $U$ contains unary operations, e.g. $U = \left\{\frac{d}{dt}, \frac{d^2}{dt^2}, \exp, \sin, \log\right\}$, $B= \{ +, -, \cdot \}$ the subset of binary expressions, $C\subset \mathbb{R}$ the subset of constants, and $A = \{ \alpha, t \}$ the subset of variables. A candidate ODE can then be constructed by sampling a walk $w$ on the graph $\mathcal{G}$. Subsequently, $\mathrm{S}$ is defined as the set of all finite walks,
 \begin{equation}
    \mathrm{S} = \{ l(w) \ | \ w = \{\mathit{v}_1, \mathit{v}_2, ..., \mathit{v}_n \} \in \mathcal{V}^*, \ (\mathit{v}_i, \mathit{v}_{i+1}) \in \mathcal{E} \ \  \forall \ \ 1 \leq i < n \},
\end{equation}
where $\mathcal{V}^* = \bigcup_{n=0}^\infty \mathcal{V}^n$ is the Kleene star of the set $\mathcal{V}$ (i.e., $\mathcal{V}^*$ is the set of all possible concatenations of sequences in $\mathcal{V}$ of different lengths, respectively in our case of labels in $L$. The set can be empty and the same labels can appear multiple times.) and $n$ is the sequence length. However, a rejection sampling strategy needs to be considered, as this approach does not necessarily produce semantically and syntactically correct expressions \cite{kissas_language_2024}. Ensuring syntactically valid expressions is a key challenge across all iterative methods. For example, dictionary-based methods achieve this naturally by restricting the search only to terms included in a predefined library \cite{brunton_discovering_2016}. Other methods consider user-specified constraints on the tree generation which restrict the choices of children given their parents \cite{cranmer_interpretable_2023}. However, these constraints are local and meaningful provided that the expression sampled is semantically valid.  

To systematically define local, between parent and children nodes, and global constraints, across subtrees, we consider a formal grammar and more specifically a context-free grammar (CFG) approach. A CFG $\mathit{G}$ is defined as the tuple $\{ \mathcal{T}, \mathcal{N}, \mathcal{P}, \mathcal{S} \}$, where 
$\mathcal{T}$ is the set of terminal symbols, $\mathcal{N}$ the set of non-terminal symbols and 
$\mathcal{T} \cap \mathcal{N} = \emptyset$, $\mathcal{P}$ a finite set of production rules, and $\mathcal{S} \in \mathcal{N}$ the starting symbol. Each rule $r \in \mathcal{P}$ is a map $\alpha \rightarrow \beta$, where $\alpha \in \mathcal{N}$ and $\beta \in (\mathcal{T} \cup \mathcal{N})^*$. A language $\mathcal{L}(\mathit{G})$ is defined as the set of all terminal strings derived by applying the production rules of the grammar starting from $\mathcal{S}$:
\begin{equation}
    \mathcal{L}(\mathit{G}) = \{ w \in \mathcal{T}^* \mid \mathcal{S} \rightarrow^* w \},
\end{equation}
where $\rightarrow^*$ implies $n_r \geq 0$ applications of rules in $\mathcal{P}$. There are two fundamental operations associated with a terminal expression, namely parsing and generation. Given an expression $w \in \mathcal{T}^*$, a parsing algorithm extracts a sequence of rules starting from $\mathcal{S}$, and given a sequence of rules, a generation algorithm derives a terminal string $w \in \mathcal{T}^*$ starting from $\mathcal{S}$. We define an interpretation map $\mathcal{I}: \mathcal{L} (\mathit{G}) \rightarrow \mathcal{O}$ \cite{oikonomou_neuro-symbolic_2025}, which assigns semantic meaning to each $w$ in terms of an ODE. The set of all ODEs represented by the grammar $\mathit{G}$ is:
\begin{equation*}
    \mathcal{O} (\mathit{G}) = \{ O_w: D \rightarrow \mathbb{R} \mid O_w = \mathcal{I}(w),  w \in \mathcal{L} (\mathit{G}) \}.
\end{equation*}

Grammar-based methods enforce validity through their rule-based structures. Additionally, domain knowledge can be incorporated to constrain the search space and further guide the discovery process \cite{kissas_language_2024}. In practice, the grammar is easily defined, via a chart parser provided by the Python package \textit{nltk} \cite{bird_natural_2009}, by only providing production rules and terms such as $\sin$ or $\exp$. An example of a simple CFG is given as follows:
\begin{align*}
    \mathcal{S} = \:& \{S\} \\
    \mathcal{N} = \:& \{S,E, V, F\} \\
    \mathcal{T} = \:& \{+, t, \sin(\cdot), \cos(\cdot), C\} \\
    \mathcal{P} = \:& \{S\rightarrow E + E\: (1), \\
    & E \rightarrow V \: (2)\:|\: F \: (3)\:|\: C \: (4), \\ 
    & F \rightarrow \sin(V) \: (5)\:|\: \cos(V) \: (6),\\
    & V \rightarrow t \: (7)\: \}, 
\end{align*}
where the expression $C+\sin(t)$ is given by the sequence $\mathbf{r} = [1,4,3,5,7]$,  $C$ is a placeholder for a constant, and $(\cdot)$ defines the rule index. For more general cases, the sequence of rules $\mathbf{r}$ is given by $\mathbf{r} = [r_1, ..., r_{n_r}]$. In practice, the length $n_r$ of the rule sequence might vary depending on the complexity of the expression (more details on the complexity metric can be found at the end of \cref{ch2:subsec:optim_obj}). In this case, the sequence $\mathbf{r}$ can be padded to a user-specified maximum length $N_{\text{max}}$ decided by the length of the most complex expression. To do so, a special padding rule $r_p \notin \mathcal{P}$ is added to the production rules $\hat{\mathcal{P}} = \mathcal{P} \ \cup \ \{ r_p\}$. Then, the padded expression is given as $\hat{\mathbf{r}} = [r_1, ..., r_{n_r}, r_p,...,r_p ]$ where $r_p$ is repeated $N_{\text{max}}-n_r$ times. 

\subsection{Manifold learning of candidate ODEs}
\label{ch2:subsec:gvae}

\begin{figure}[h!]
    \centering
    \includegraphics[scale=0.85]{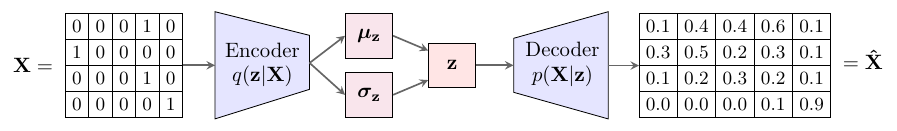}
    \caption{Overview of the GVAE which encodes and decodes a sequence of rules.}
    \label{fig:ch2:gvae}
\end{figure}

As searching a very large discrete space is inefficient \cite{brence_probabilistic_2021}, we propose a different approach where we first embed the discrete expressions into a continuous low-dimensional space which we then search. For embedding the expressions sampled from the grammar, we consider the GVAE, an adaptation of the classic variational autoencoder \cite{kingma_auto-encoding_2014} proposed by Kusner et al. \cite{kusner_grammar_2017}. The architecture is based on the inherent operations of parsing and generating that a formal grammar possesses. The encoder first parses an expression to yield a sequence of rules, which is subsequently transformed into one-hot encoded vectors. A different function accepts the matrix of one-hot encoded vectors $\mathbf{X}$ as input and maps it to a latent distribution $q(\mathbf{z}|\mathbf{X})$, see \cref{fig:ch2:gvae}. The decoder then maps the reparametrized latent vector $\mathbf{z}$ to a sequence of rules through $p(\mathbf{X}|\mathbf{z})$, see Kusner et. al. \cite{kusner_grammar_2017} for more details. 

One drawback of CFGs is the way they represent real numbers using compositions of structural non-terminals, such as dots, and integer digits. For example, real numbers are constructed using the following subgrammar: 
\begin{align*}
    \mathcal{S} = \:& \{C\} \\
    \mathcal{N} = \:& \{C, D, N\} \\
    \mathcal{T} = \:& \{., 0,1,2,3,4,5,6,7,8,9\} \\
    \mathcal{P} = \:& \{C \rightarrow N.N \: (1), \\
    & N \rightarrow D \: (2) \: | \: ND \: (3),\\
    & D \rightarrow 0 \: (4) \: | \: 1 \: (5) \: | \: 2 \: (6) \: | \: 3 \: (7) \: | \: 4 \: (8) \: | \: 5 \: (9) \: | \: 6 \: (10) \: | \: 7 \: (11) \: | \: 8 \: (12) \: | \: 9 \: (13) \}.
\end{align*}
Therefore, the number $1.234$ is constructed using the following process:
\begin{align*}
    C &\xrightarrow{(1)} N.N, \\
    &\xrightarrow{2\times (3)} N.NDD, \\
    &\xrightarrow{2\times (2)} D.DDD, \\
    &\xrightarrow{(5), (6), (7), (8)} 1.234,
\end{align*}
which results in nine rule applications for a number with precision three. This representation of numbers thus leads to very long sequences of rules for high-precision arithmetic. For this reason, we refrain from encoding numbers, as proposed in the literature \cite{kissas_language_2024,oikonomou_neuro-symbolic_2025,kusner_grammar_2017}, but instead embed skeletons of expressions and then optimize for the constants in a two-step approach; details can be found in the next subsection. 

The encoder of the one-hot encoded vectors is made of convolutional neural networks. Typically, we use three layers with the  rectified linear unit (ReLU) activation function, where the dimensions of the input, the output, and kernel size depend on the specific problem and complexity of the grammar. We adopt recurrent neural networks for the decoder function, namely gated recurrent units (GRU) layers \cite{cho_learning_2014}, with the activation function comprising exponential linear units (ELU) \cite{clevert_fast_2015}.
We train the GVAE model through the loss:
\begin{align}
    \mathcal{L} & = \mathcal{L}_{BCE} + \upbeta_{KL} \mathcal{L}_{KL} , \label{eq:train_loss} 
\end{align}
where $\mathcal{L}_{BCE}$ is the binary cross-entropy loss with a Sigmoid layer between the baseline $\mathbf{X}$ and predicted $\mathbf{\hat{X}}$ one-hot encoded vectors, $\mathcal{L}_{KL}$ the Kullback-Leibler divergence loss, and $\upbeta_{KL}$ a weight factor to balance the different loss objectives. We use the Adam optimizer \cite{kingma_adam_2014}, a stochastic gradient-based optimization method that adapts its estimates of lower-order moments, with an initial learning rate of $0.001$ and an adaptive learning rate scheduler, reducing the learning rate if the metric has stopped improving. The multiplicative factor is chosen as $0.9$ and the remaining hyperparameters are defined for each model individually.

\subsection{Latent space optimization}
\label{ch2:subsec:latent_optim}

As discussed above, the GVAE is first pre-trained using symbolic expressions of ODEs. Then, the latent space is searched for the ODE that satisfies a given numerical trajectory in two stages. For the outer loop of the optimization procedure (first stage), we consider the Covariance Matrix Adaptive Evolution Strategy (CMA-ES) \cite{hansen_cma_2016, hansen_completely_2001,hansen_reducing_2003,hansen_cma-espycma_2024}; a non-linear non-convex gradient-free optimization scheme. CMA-ES samples a population from a multivariate normal distribution, whose mean and covariance matrix are regularly updated and adapted through a minimization objective, making it particularly suitable for complex loss landscapes. For the inner loop (second stage), we consider the optimization method proposed by Nelder et. al. \cite{nelder_simplex_1965, virtanen_scipy_2020} for adapting the constants. 

We assume noisy and sparse observations, where sparsity refers to not having all types of measurements of the trajectory and solution derivatives $u_{t_i}$, $\dot{u}_{t_i}$, or $\ddot{u}_{t_i}$ at times $t_i$ for $i=0..n_s-1$, where $n_s$ is the number of measurements. Due to the noise and sparsity, we need to infer the missing information, meaning either integrate or differentiate to obtain the missing mode. Typically, finite difference schemes can be used here, which, however, struggle with noise and require sophisticated and individually tuned filtering schemes. For this reason, we consider a combination of a multilayer perceptron (MLP) with three layers, 32 neurons, and Sigmoid linear unit (SiLU) activation \cite{elfwing_sigmoid-weighted_2017} to smoothen the data ($u^{NN}_t$) and automatic differentiation to approximate the derivatives ($\dot{u}^{NN}_t, \ddot{u}^{NN}_t$) \cite{paszke_automatic_2017}. 

\subsection{Optimization objectives}
\label{ch2:subsec:optim_obj}

For evaluating how well a given trajectory satisfies an ODE, we need to perform symbolic operations, for which we use \textit{SymEngine} \cite{symengine_development_team_symengine_2023}, a fast symbolic manipulation library in C++, accessible through a Python wrapper, and the Python package \textit{SymPy} \cite{meurer_sympy_2017}. Two types of objectives are introduced: the differential equation residual loss $\mathcal{L}_{DE}$ and the standardized mean squared error of the solution trajectories $\mathcal{L}_{SOL}$:
\begin{align}
    \mathcal{L}_{DE}(\tilde{u}_t) & = \sqrt{\frac{1}{n_s}\sum_{i=0}^{n_s-1}\left(\mathcal{D}\left(u^{NN}_{t_i}, \dot{u}^{NN}_{t_i}, \ddot{u}^{NN}_{t_i}\right)-F({t_i})\right)^2} \label{eq:optim_DE_loss} \\
    \mathcal{L}_{SOL}(\tilde{u}_t, \hat{u}_t) & = \frac{\frac{1}{n_s}\sum_{i=0}^{n_s-1}(u^{NN}_{t_i}-\hat{u}_{t_i})^2}{\frac{1}{n_s}\sum_{i=0}^{n_s-1}{u^{{NN}^2}_{t_i}}} +\frac{\frac{1}{n_s}\sum_{i=0}^{n_s-1}(\dot{u}^{NN}_{t_i}-\hat{\dot{u}}_{t_i})^2}{\frac{1}{n_s}\sum_{i=0}^{n_s-1}{\dot{u}^{{NN}^2}_{t_i}}}+\frac{\frac{1}{n_s}\sum_{i=0}^{n_s-1}(\ddot{u}^{NN}_{t_i}-\hat{\ddot{u}}_{t_i})^2}{\frac{1}{n_s}\sum_{i=0}^{n_s-1}{\ddot{u}^{{NN}^2}_{t_i}}},
    \label{eq:optim_SOL_loss}
\end{align}
where $\mathcal{D}$ is the differential operator, $u(t)$ the dependent function, $t$ the independent variable, and $F(t)$ the force term. Therefore, the ODE can be represented by the implicit form $\mathcal{D}(u(t))-F(t)=0$. Solving the ODE can be computationally taxing; hence, the ODE is not solved for every equation, where $\hat{u}_t$, $\hat{\dot{u}}_{t}$, and $\hat{\ddot{u}}_{t}$ are the solution trajectories of the predicted ODE. First, the residual loss $\mathcal{L}_{DE}$ is determined, and if this loss is below a certain threshold $\theta_{DE}$, then $\mathcal{L}_{SOL}$ is evaluated. During the optimization process, we need to ensure that we are comparing the same equations with each other regardless of scaling to avoid ill-posedness. We normalize our sampled equation by its highest derivative so that such expressions have the same loss $\mathcal{L}_{DE}$. In cases where solving the ODE is discarded or omitted, it needs to be prevented that GODE generates trivial or close-to-trivial equations, such as \cref{eq:trivial1,eq:trivial2}, which would yield small losses irrespective of the dependent functions. In fact, symbolic regression methods are commonly defined as function approximators, meaning that they attempt to discover a functional mapping $f(X(t))$ from an input $X(t)$ to an output $y(t)$: $y(t) = f(X(t)), t\in \mathbb{D} $, where $t$ is the independent variable, often representing time, over the domain $\mathbb{D}$. This problem is typically referred to as solution discovery \cite{hyeonjung_survey_2023}. However, fewer works have tackled the equation discovery problem, i.e., discovering differential equations directly, mostly due to the inherent ill-posedness of such a task (owing to noise, sparsity, or nonuniqueness) \cite{scholl_symbolic_2022}. As aforementioned, dictionary-based methods are often adopted to this end, focusing on linear regression over candidate functions \cite{brunton_discovering_2016}, while other schemes attempt to reformulate and reduce the problem to explicit ODEs \cite{dascoli_odeformer_2023,cranmer_interpretable_2023}. The latter seek to infer this explicit (or also known as functional) form of $\dot{X}(t)$ as $\dot{X}(t) = f(X(t))$ over a specified domain $\mathbb{D}$. Instead, we propose to directly generate implicit equations (i.e., for this case $f(X(t),\dot{X}(t)) = 0$ or for ODEs $\mathcal{D}(u(t))-F(t) = 0$), which are more expressive than explicit ODEs. However, a common challenge faced in the discovery of implicit equations \cite{riolo_symbolic_2010} is encountering trivial solutions:
\begin{align}
    \frac{du(t)}{dt}+2-\frac{du(t)}{dt}-2 & = 0 \label{eq:trivial1}\\
    \frac{d^2u(t)}{dt^2}-0.985\frac{d^2u(t)}{dt^2} & \approx 0. \label{eq:trivial2}
\end{align}
Hence, we filter these cases out by evaluating the equations for a range of different possible solutions $\mathcal{S}^*$. 
\begin{align}
    \frac{1}{n\left(\mathcal{S}^*\right)}\sum_{u_{\mathcal{S}^*}\in\mathcal{S}^*}\left(\frac{1}{n_s}\sum_{i=0}^{n_s-1} \left\|\mathcal{D}(u_{\mathcal{S}^*}(t_i))-F({t_i})\right\| \right)&\leq\varepsilon\label{eq:trivial_sol}
\end{align}
If the mean of all mean evaluations are below a certain threshold $\varepsilon$, we treat the equation as trivial and discard it; respectively return a high penalty. 

As discovered ODEs should also be parsimonious, a new evaluation metric was introduced for the search problem of some examples, a variation of typical model selection criteria:
\begin{equation}
    \mathcal{L}_{IC} = \alpha \mathcal{C}+(1-\alpha)n_s\mathcal{L}_\text{accuracy} \label{eq:method_ic}
\end{equation}
where $\alpha$ is a weight factor, $\mathcal{L}_\text{accuracy}$ the loss of the ODE evaluation, and $\mathcal{C}$ the complexity metric. The complexity metric $\mathcal{C}$ is assessed by evaluating the attributes of an expression and counting the number of constants, variables, and operations, where each is assumed to have a weight of $1$. However, due to the automated extraction of the attributes with \textit{SymPy}, certain operations are weighted more. For instance, the ODE $5\frac{d}{dt}u(t)+25u(t)-\sin(t)=0$ has a complexity of 16, where alone the derivatives such as $\frac{d}{dt}u(t)$ or $\frac{d^2}{dt^2}u(t)$ count as 6 and the function $u(t)$ as 2. Automating this process requires significant simplifications that do not necessarily need to be included in typical simplification functions, such as in the one employed by \textit{SymPy}. To reduce this bias, the weight factor $\alpha = 0.1$ is chosen to be fairly small. 
\section{Results: data generation, training, and discovery}
\label{ch3:results}

We present three types of benchmarks to evaluate GODE. We focus on well-designed synthetic examples to thoroughly test the potency of the employed approach, which can then be used in conjunction with experimental data. The first benchmark (\cref{{ch3:subsec:bench1}}) involves comparing GODE with three state-of-the-art methods: ODEFormer \cite{dascoli_odeformer_2023}, PySR \cite{cranmer_interpretable_2023}, and ProGED \cite{brence_probabilistic_2021,omejc_probabilistic_2024}, on one-dimensional explicit ODEs. ODEFormer is a transformer-based direct algorithm and follows the scheme described in \cref{ch1:intro}: A large deep learning model is trained to accept numerical trajectories. Subsequently, the model embeds these with tokenization, passes the tokens through a transformer, and decodes them to predict mathematical symbolic ODEs. Such an approach learns patterns and structure through data and, hence, requires a large dataset and rejection sampling to ensure syntactic validity. In contrast, PySR is a popular genetic programming-based discrete search tool, which exploits parallelization and a multi-population evolutionary algorithm based on tournament selection to discover functions computationally efficiently. A previous study by d'Ascoli et al. \cite{dascoli_odeformer_2023} demonstrated that ODEFormer performs on par or better than PySR justifying the selection of these two methods. ProGED uses formal grammars, specifically probabilistic CFGs, to infer symbolic expressions with Monte-Carlo sampling. Omejc et al. \cite{omejc_probabilistic_2024} have recently improved its methodology and, thus, ProGED is included as a reference model. We do not compare with dictionary-based methods, such as Sparse Identification of Nonlinear Dynamics (SINDy), due to its main disadvantage of requiring a preselection of basis functions, which depends on domain knowledge (see \cref{ch1:intro}). Moreover, if we relax the constraint of the highly specific candidate expressions, then the main challenge becomes searching immensely high-dimensional spaces. The accuracy and generality of the search is the motivation of this comparison. One-dimensional explicit ODEs are selected because all three reference models require explicit forms of ODEs. Notably, PySR, originally designed for symbolic regression between dependent and independent variables (i.e., function approximations), can be adapted to seek explicit (see \cref{ch2:subsec:optim_obj}) ODEs. Although ProGED can technically be adapted to accept implicit ODEs with a suitable grammar template and adaptations in the evaluation of candidate expressions, such an adaptation lies outside the scope of this work. We here focus on explicit forms for consistency. Lastly, only ODEFormer requires providing time series trajectories of the respective ODEs for its training, which demands solving these ODEs. As this can be computationally costly, we assess the sample and parameter efficiency of ODEFormer by also training smaller models. 

In the second benchmark (\cref{ch3:subsec:bench2}), we train the proposed model to handle both first- and second-order linear or nonlinear ODEs that can also be implicit. The third benchmark (\cref{ch3:subsec:bench3}) aims at discovering three common second-order ODEs from nonlinear dynamics based on simulated noisy acceleration measurements. For both the second and third benchmarks, we compare the performance of our GODE with that of PySR and ProGED, as ODEFormer did not perform satisfactorily for the first benchmark.

To identify trivial ODEs, we define the set of possible solutions $\mathcal{S}^*$ as $\left\{t, -2.5t, \sin(3t),2\cos\left(\frac{t}{4}\right)+\frac{t}{3}\right\}$. We run all methods for five independent runs for each example to minimize the effect of random initializations. To facilitate the understanding of each method, details regarding the input and output expectations are provided in \cref{app:problem_form_method}. 

\subsection{One-dimensional explicit ODEs}
\label{ch3:subsec:bench1}
This comparison focuses on one-dimensional explicit ODEs to evaluate GODE against three state-of-the-art models: ODEFormer \cite{dascoli_odeformer_2023}, ProGED \cite{omejc_probabilistic_2024,brence_probabilistic_2021}, and PySR \cite{cranmer_interpretable_2023}, all of which only support explicit forms of ODEs in their implementations. Furthermore, we trained smaller versions of ODEFormer to assess its performance with fewer parameters and reduced computational resources on a smaller training dataset. All four types of models are tested on 30 different one-dimensional ODEs, of which 23 are taken from the ODEBench provided by d'Ascoli et al. \cite{dascoli_odeformer_2023} (see \cref{tab:app1:odebench,tab:app1:add}).

\subsubsection{Library generation}
\label{ch3:subsubsec:b1_lib}
The grammar for the first benchmark comprises 29 production rules, see \cref{app:Bench1}, with an assumed maximum expression length of $N_\text{max}=40$. We generated 10,000 skeletons of expressions and split them into a 9:1 ratio for training and testing of the GVAE. Generating the dataset for training smaller models of ODEFormer, comprising 15,000 skeleton expressions using only the grammar, took approximately one to two minutes on a MacBook Pro M3 (Apple M3 Pro, 10-core CPU, 18-core GPU, and 36GB unified memory). The time integration of 70,000 expressions (five random sets of constants per skeleton expression), employing the filtering scheme described in Ascoli et al. \cite{dascoli_odeformer_2023}, required about five hours. This dataset is fairly small compared to the training dataset of the original ODEFormer model, which contained $50$ million examples.


\subsubsection{Training and inference}
\label{ch3:subsubsec:b1_train}
Generally, for the GVAE model, some hyperparameters, such as the hidden dimension of the GRU layers, are kept as default, whereas others, such as the latent dimension, were manually tuned. For this model, we selected a latent dimension of 24, kernel sizes 7, 8, and 9 for the three convolutional neural network layers of the encoder, and a hidden dimension of 80 for the three bidirectional GRU layers of the decoder. The weight factor of the loss function was set to $\upbeta_{KL}=10^{-3}$, and early stopping was applied after 1,000 epochs. The learning rate scheduler used a patience of 400 epochs and a minimum learning rate of 0.00005. Training a single model on one GPU (NVIDIA GeForce RTX 4090) took approximately 10 to 20 minutes. Each generated expression represents $f(u^{NN}_t)$, to which $-\dot{u}^{NN}_t$ was added to form first-order ODEs. This addition addresses the ill-posedness of implicit ODEs as the coefficient of the first derivative is consistently 1. We employed CMA-ES, as described in \cref{ch2:subsec:latent_optim}, with a population size of 100, 10 generations, and an initial standard deviation of 0.5, to identify the most optimal ODE. Since this benchmark involves only first-order ODEs, the latent space optimization focused on minimizing the ODE residual loss $\mathcal{L}_{DE}$ without solving the ODE.

For ODEFormer, we employed the original code available on GitHub, and trained it under two different configurations. The dimensions of both the encoder and decoder were reduced to either 64 or 128, with the option to re-scale the input enabled. Each model was trained for 24 hours on a single GPU, achieving about 100,000 steps. During inference, ODEFormer applied beam search with a beam size of 50 and beam temperature of 0.1, which are the default parameters from the original paper \cite{dascoli_odeformer_2023}, and uses the metric $R_2$ for selection.

PySR, which requires no training as it uses an evolutionary algorithm, was used directly from the Python package (version 0.19.4). For the inference task, the model selection option was set to \verb|'best'|, selecting the equation that best balances accuracy and complexity. The number of populations was set to $20$, the population size to $30$, and the number of iterations to $20$ \cite{cranmer_interpretable_2023}. PySR allows defining a loss between target and prediction, for which we chose the squared error.

For ProGED (version 0.8.5), we opted for the provided grammar \verb|'universal'|, a sample size of $100$, and a maximum number of repetitions of $100$. ProGED evaluated the root-mean-squared error of the solution trajectories of the predicted equation \cite{omejc_probabilistic_2024, brence_probabilistic_2021}.

\subsubsection{Evaluation}
\label{ch3:subsubsec:b1_eval}

Each example was evaluated by sampling the time series within a domain, using a sampling frequency that results in 50 to 300 evaluation points, and applying predefined initial conditions. Each example was then corrupted with 5\% Gaussian noise in order to account for the setting where actual data are available, which are expected to be affected by noise (e.g., from measurement devices). For ODEFormer, only the time series $\tilde{u}_t$ was provided, whereas PySR, ProGED, and GODE required an approximation of the derivative $\dot{u}^{NN}_t$. To obtain this approximation, an MLP (\cref{ch2:subsec:latent_optim}) was trained on the noisy observations of $\tilde{u}_t$ over 10,000 epochs with a batch size of $32$, utilizing the Adam optimizer \cite{kingma_adam_2014}. A stepped learning rate scheduler was used, starting at a learning rate of $0.001$ with a step size of $500$ and a multiplicative factor of the learning decay rate of $0.95$ \cite{ansel_pytorch_2024}. The relative L2 error, which compares the solutions of the predicted ODE $\hat{\boldsymbol{u}}$ with the ground truth trajectories $\boldsymbol{u}$, is defined as follows:
\begin{equation}
    \text{Relative L2 error}(\boldsymbol{u},\hat{\boldsymbol{u}}) = \frac{\|\boldsymbol{u}-\hat{\boldsymbol{u}}\|_2}{\|\boldsymbol{u}\|_2} \label{eq:l2}
\end{equation}

\begin{figure}[h!]
    \centering
    \includegraphics[scale=0.9]{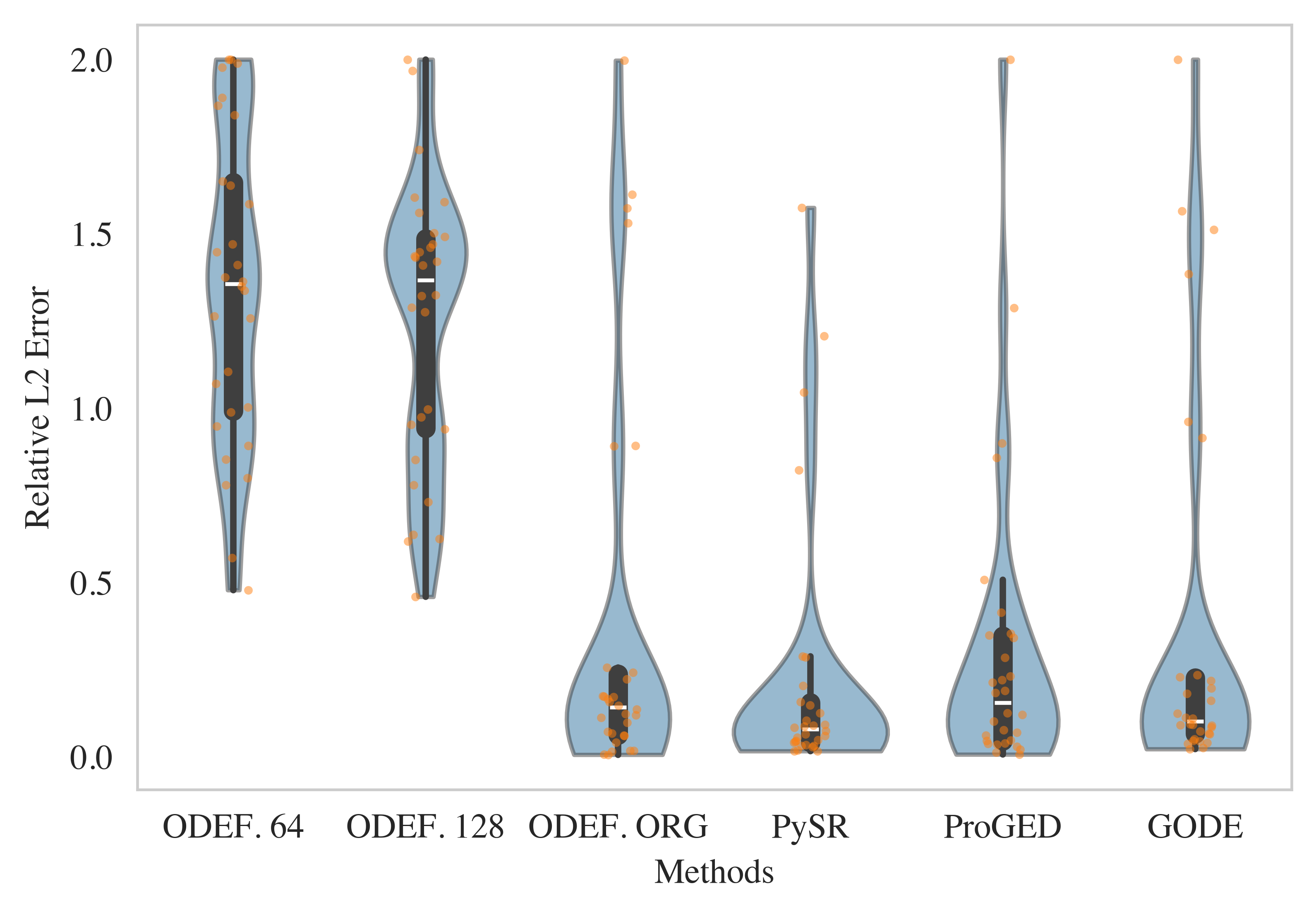} 
    \caption{Violin plots of the relative L2 errors for the predicted time series based on the predicted equations with individual errors marked as orange dots for each investigated method.}
    \label{fig:b1_res_violin_l2}
\end{figure}

\begin{figure*}[h!]
    \centering
    \includegraphics[width=\linewidth]{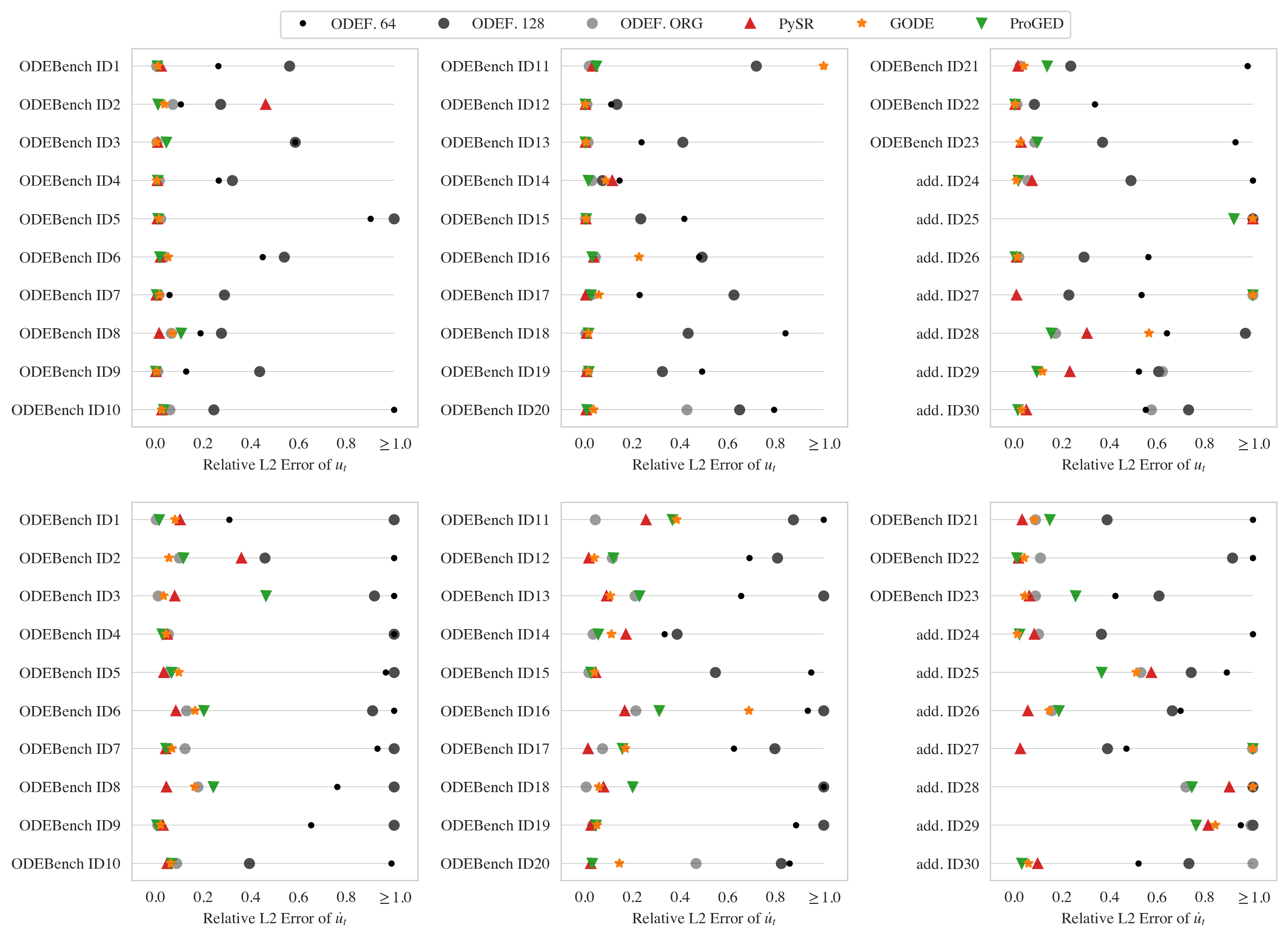}
    \caption{Relative L2 errors for the predicted time series based on the predicted equations of each example of the first benchmark.}
    \label{fig:b1_detailed}
\end{figure*}

\begin{table}[b!]
    \centering
    \caption{Mean relative L2 errors of $u_t$ and $\dot{u}_t$ for different models for the first benchmark. If the relative L2 error is above 1, it was floored to 1.}
    \renewcommand{\arraystretch}{1.5}
    \begin{tabular}{|p{3.3cm}|p{1.2cm}|p{1.3cm}|p{1.3cm}|}
    \hline
    Model & L2 $>$ 1 & \multicolumn{2}{|c|}{Mean Relative L2 Error}  \\
    \cline{3-4}
    & & $u_t$  &   $\dot{u}_t$ \\ 
    \hline
    \hline
    ODEFormer 64 & 13 & 0.524 & 0.816 \\
    ODEFormer 128 & 13 & 0.454 & 0.790  \\
    ODEFormer ORG & 2 & 0.148 & 0.224  \\
    PySR best & \textbf{1} & \textbf{0.084} & \textbf{0.148} \\
    ProGED & 2 & 0.095 & 0.211 \\
    GODE (ours) & 5 & 0.150 & 0.211  \\
    \hline
    \end{tabular}
    \label{tab:b1_l2}
\end{table}

\cref{fig:b1_res_violin_l2} presents violin plots that summarize the relative L2 errors for $u_t$ and $\dot{u}_t$ across each method from 30 examples (\cref{app:Bench1}). The violin plots show the distribution of the errors and further include a rotated kernel density plot on each side. This gives a sense of the spread of the computed relative error and an indication on where values are more or less concentrated (wider portion of the plot). To prevent outliers from skewing the statistics too excessively, errors exceeding 1.0 were rounded down to 1.0. Furthermore, the mean relative L2 errors are listed in \cref{tab:b1_l2}. \cref{fig:b1_detailed} shows the predicted results in terms of the relative L2 error for $u_t$ and $\dot{u}_t$, separately for all six tested models (ODEFormer 64, ODEFormer 128, ODEFormer ORG, PySR, ProGED, and GODE). Both ODEFormer models with fewer parameters (ODEFormer 64: 3.0 million and ODEFormer 128: 6.9 million vs. ODEFormer ORG: 60.7 million) and smaller training datasets perform significantly worse than the original ODEFormer model, as well as PySR, ProGED, and our model GODE. These reduced models tend to predict more inaccurate solutions compared to the others. Moreover, ODEFormer does not ensure validity and might predict expressions such as $\log(-u(t))$. Our model performs better than the original ODEFormer model, despite the latter having over 190 times more parameters, being trained on a much larger dataset (around a 1,000 times more samples), and accessing more computational resources (three days of training on a single NVIDIA A100 GPU with 80GB memory and 8 CPU cores) \cite{dascoli_odeformer_2023}. Finally, while our model performs slightly worse than PySR and ProGED, PySR appears to predict the most accurate predictions. Due to the computationally intensive data generation and training of ODEFormer, the difficulty of adapting it to new tasks, and the overall average performance, the remaining benchmarks only include comparisons with PySR and ProGED.

\subsection{Linear and nonlinear ODEs}
\label{ch3:subsec:bench2}
The second benchmark involves discovering implicit linear and nonlinear first- or second-order ODEs. Given that the force term $F(t)$ is unknown, there can be a multitude of different tuples of the differential operator $\mathcal{D}(u(t))$ and $F(t)$ that fit a single set of solution trajectories. The benchmark dataset comprises five linear and five nonlinear ODEs with known analytical solutions, many of which are taken from the seminal work of Tsoulos and Lagaris \cite{tsoulos_solving_2006} (see \cref{tab:app2:bench2}). 

\subsubsection{Library generation, training, and inference}
\label{ch3:subsubsec:b2_train}
For this benchmark, two different grammars were employed: one for the dataset generation, featuring 29 production rules and a maximum expression length of 50, and another for the GVAE, comprising 24 rules with a maximum length of 70, see \cref{app:Bench2}. The grammar for the GVAE is less specific to give more freedom to the model to learn the structure, while we want to control recursions during the dataset generation better. During the training of the GVAE, 50,000 skeleton expressions were generated, with an additional 1,000 reserved for testing.

For this GVAE model, we selected a latent dimension 26, a kernel size of 7 for all three convolutional neural network layers of the encoder, and a hidden dimension of 80 for the three bidirectional GRU layers of the decoder. The weight factor for the loss function was set to $\upbeta_{KL}=10^{-4}$ and early stopping was applied after 1,000 epochs. The learning rate scheduler used a patience of 500 and a minimum learning rate of 0.0001. To reduce ill-posedness during optimization, the first coefficient of the symbolic skeleton expression was fixed at $1$. For latent space optimization, the CMA-ES algorithm was configured with a population size of 200, 10 iterations, and an initial standard deviation of 0.5. The initial threshold for the change in ODE evaluation was set to $\theta_{DE}=200$, and for the subsequent iterations, it was dynamically adjusted to the mean of the 20 lowest losses of the current iteration with a 5\% buffer margin. Furthermore, the objective during the search was $\mathcal{L}_{IC}$ (\cref{eq:method_ic}), which balances accuracy and sparsity of candidate ODEs.

For both ProGED and PySR, the problem was reformulated to fit the explicit form of ODEs. This constraint might lead to inaccuracies in predicting implicit ODEs, particularly ODEs which cannot be written in the explicit form. The settings from the first benchmark were retained for both models (\cref{ch3:subsubsec:b1_train}). In addition, for a fair comparison, the model selection option \verb|'score'| was also evaluated for PySR.

\subsubsection{Evaluation}
\label{ch3:subsubsec:b2_eval}
The evaluation process follows the same protocol as the previous benchmark, see \cref{ch3:subsubsec:b1_eval}, with additionally the specification of the order of the expected ODE. When seeking a second-order ODE, autodifferentiation is applied twice to the trained MLP to approximate the second derivative $\ddot{u}_t$. 

\begin{figure*}[t!]
    \centering
    \includegraphics[width=\textwidth]{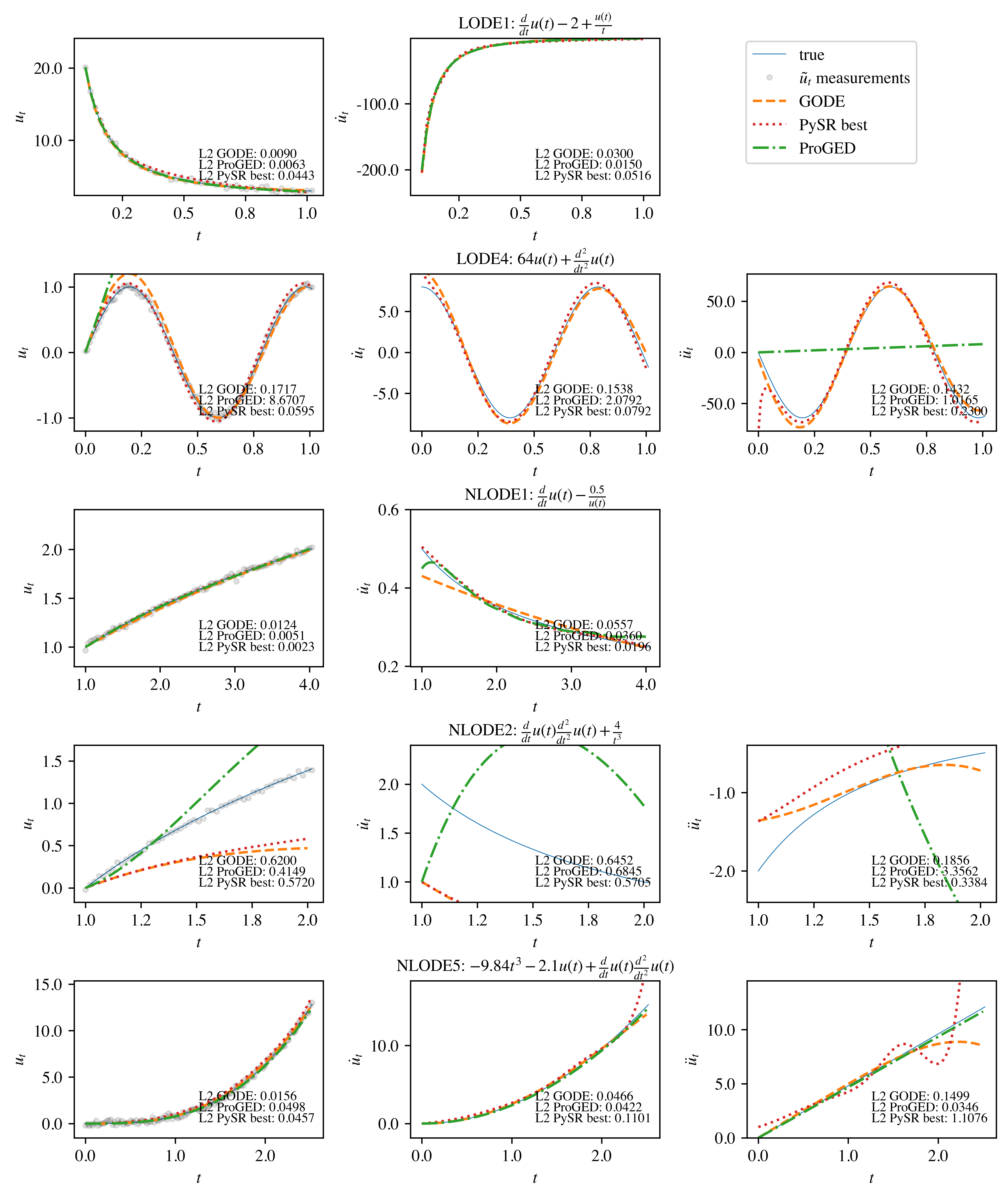}
    \caption{Solution trajectories of the predicted and ground truth ODEs of five selected examples from the second benchmark. The remaining examples can be found in \cref{app:Bench2}.}
    \label{fig:b2_res}
\end{figure*}

\cref{fig:b2_res} displays the solution trajectories of the predicted and ground-truth ODEs of five selected ODEs, alongside the relative L2 error of $u_t$, $\dot{u}_t$, and $\ddot{u}_t$ for the methods: ProGED, PySR best, and our GODE. The plots of the remaining ODEs are shown in \cref{fig:app2:bench2} in \cref{app:Bench2}. Our model successfully predicts implicit ODEs. Nonetheless, certain cases, such as NLODE2, prove to be challenging for all models, potentially due to the noisy data and hence difficulty in approximating the solution trajectories with the MLP. Generally, PySR performs well, even for differential equations beyond its exploration space (i.e., implicit ODEs, which cannot be formulated into the explicit form). However, all models, but particularly in the cases of PySR and ProGED, might discover ODEs, which can be both formulated in the implicit and explicit forms, whose solution trajectories align well with the noisy data, thus approximating the functions effectively. Nonetheless, there is no guarantee that the discovered equation mimics the desired dynamical characteristics, as multiple different ODEs could fit a single solution set. The model selection choice \verb|'score'| shows slightly lower accuracy compared to \verb|'best'| but utilizes significantly simpler expressions. The performance of ProGED varies; it excels in cases such as LODE1 or NLODE5, but performs very poorly in LODE4 or NLODE4, possibly due to its limited grammar template or unsuitability for implicit ODEs. Our GODE outperforms all other methods in both accuracy and complexity, reinforcing the effectiveness of the introduced model selection criteria $\mathcal{L}_{IC}$ in \cref{eq:method_ic}.

\begin{table*}[t!]
    \centering
    \caption{Mean relative L2 errors of $u_t$, $\dot{u}_t$ and $\ddot{u}_t$ for the second benchmark. If the relative L2 error is above 1, it was floored to 1.}
    \renewcommand{\arraystretch}{1.5}
    \begin{tabular}{|p{2.5cm}|p{1.2cm}|p{1.0cm}|p{1.0cm}|p{1.0cm}|p{2.2cm}|}
    \hline
    Model & L2 $>$ 1 & \multicolumn{3}{|l|}{Mean Relative L2 Error} & Mean Relative \\
    \cline{3-5}
    & & $u_t$  &   $\dot{u}_t$&   $\ddot{u}_t$ & Complexity\\ 
    \hline
    \hline
    True & - & - & - & - & 1.0 \\
    \hline
    ProGED & 10 & 0.326 & 0.406 & 0.811 & 1.01\\
    PySR best & 1 & \textbf{0.115} & 0.145 & 0.379 & 1.74\\
    PySR score & \textbf{0} & 0.143 & 0.136 & 0.363 & 1.04 \\
    GODE (ours) & \textbf{0} & 0.120 & \textbf{0.133} & \textbf{0.187}  & \textbf{0.90}\\
    \hline
    \end{tabular}
    \label{tab:b2_l2}
\end{table*}

\subsection{Nonlinear dynamics ODEs}
\label{ch3:subsec:bench3}
The third benchmark evaluates the proposed GODE against the state-of-the-art methods ProGED \cite{omejc_probabilistic_2024,brence_probabilistic_2021} and PySR \cite{cranmer_interpretable_2023} using three engineering examples: the damped pendulum, the Duffing oscillator, and the Van der Pol oscillator. The pendulum serves as a classic example of a linear ODE in mechanics, showcasing the relationship between stiffness, damping, mass, and force through its motion. The Duffing oscillator extends this by modeling damped and driven oscillators \cite{duffing_ingenieur_1918}, whereas the Van der Pol oscillator incorporates nonlinear damping \cite{van_der_pol_lxxxviii_1926}. Typically, in an engineering setting, acceleration can be measured with accelerometers, and the force (actuation) term might be known in certain contexts (e.g., experimental testing, earthquake inputs). This benchmark assumes noisy acceleration measurements and a known (input) force term, representing a case where partial information about the sought ODE is available via input-output monitoring information.

\subsubsection{Library generation, training, and inference}
\label{ch3:subsubsec:b3_train}
In this benchmark, the CFG for the dataset generation featured 26 production rules with a maximum expression length of $N_\text{max}=40$, while the grammar for the GVAE included 22 rules and a maximum length of 65. The training dataset consisted of 40,000 generated samples, with an additional 1,000 samples reserved for testing.

For our GVAE model, the following hyperparameters were selected: a latent dimension of 21, kernel sizes of 7, 8, and 9 for the three convolutional neural network layers in the encoder, and a hidden dimension of 80 for the three bidirectional GRU layers of the decoder. The weight factor for the loss function was $\upbeta_{KL}=10^{-4}$ and early stopping was applied after 1,000 epochs. The learning rate scheduler used a patience of 500 and a minimum learning rate of 0.0001. To explore the latent space, the CMA-ES algorithm was set to a population size of 500, 5 generations, and an initial standard deviation of 0.5. The same adaptive threshold for the loss evaluation and model selection criterion from the previous benchmark were also applied here (see \cref{ch3:subsubsec:b2_train}).

For both ProGED and PySR, the problem was reformulated to fit the explicit form of ODEs. In this case, the force term cannot be induced because the coefficient of the highest derivative in the explicit form is $1$, which however does not need to be the corresponding coefficient for the induced force term. Both models used the same settings as in the first benchmark (\cref{ch3:subsubsec:b1_train}). 

\subsubsection{Evaluation}
\label{ch3:subsubsec:b3_eval}
To train the MLP, the loss needs to be adapted to approximate the second derivative $\ddot{u}_t$. However, due to approximating the MLP on second derivatives, drifts in $u_t$ and $\dot{u}_t$ might appear. Therefore, the MLP loss was adapted to:
\begin{align}
    \mathcal{L}_\text{MLP} = \:& \underbrace{\frac{1}{n_s}\sum_{i=0}^{n_s-1}\left(\ddot{\tilde{u}}_{t_i}-\ddot{u}_{t_i}^{NN}\right)^2}_{\text{prediction loss}}+\beta_{\ddot{u}}\underbrace{\frac{1}{n_s-i_T}\sum_{i=i_T}^{n_s-1}\left(\left(u_{t_i}^{NN}-{u_{t_i}^{NN\text{, detrend}}}\right)^2+\left(\dot{u}_{t_i}^{NN}-{\dot{u}_{t_i}^{NN\text{, detrend}}}\right)^2\right)}_{\text{anti-drift loss}} \notag \\
    & +\underbrace{\left(u_{t_0}-u({t_0})\right)^2+\left(\dot{u}_{t_0}-\dot{u}({t_0})\right)^2}_{\text{initial conditions loss}}, \label{eq:b3_loss_mlp}
\end{align}
where the weight coefficient $\beta_{\ddot{u}}$ was set to $10^{-3}$, $i_T$ is the index of $t_i$ at $t_{i_T}=T$, and $T$ the period of the force term. ${u_{t}^{NN\text{, detrend}}}$ and ${\dot{u}_{t}^{NN\text{, detrend}}}$ are the detrended approximated responses, where for $u_t$ a linear trend and for $\dot{u}_t$ a constant trend was removed. Moreover, the known initial conditions $u(t_0)$ and $\dot{u}(t_0)$ were added to the loss. Given the higher complexity of the solution trajectories and longer time span, the stepped learning rate scheduler with the settings in \cref{ch3:subsubsec:b1_train} was modified to a step size of $2,000$ and the MLP was trained for $100,000$ epochs. 

\begin{figure*}[h!]
    \centering
    \includegraphics[width=\textwidth]{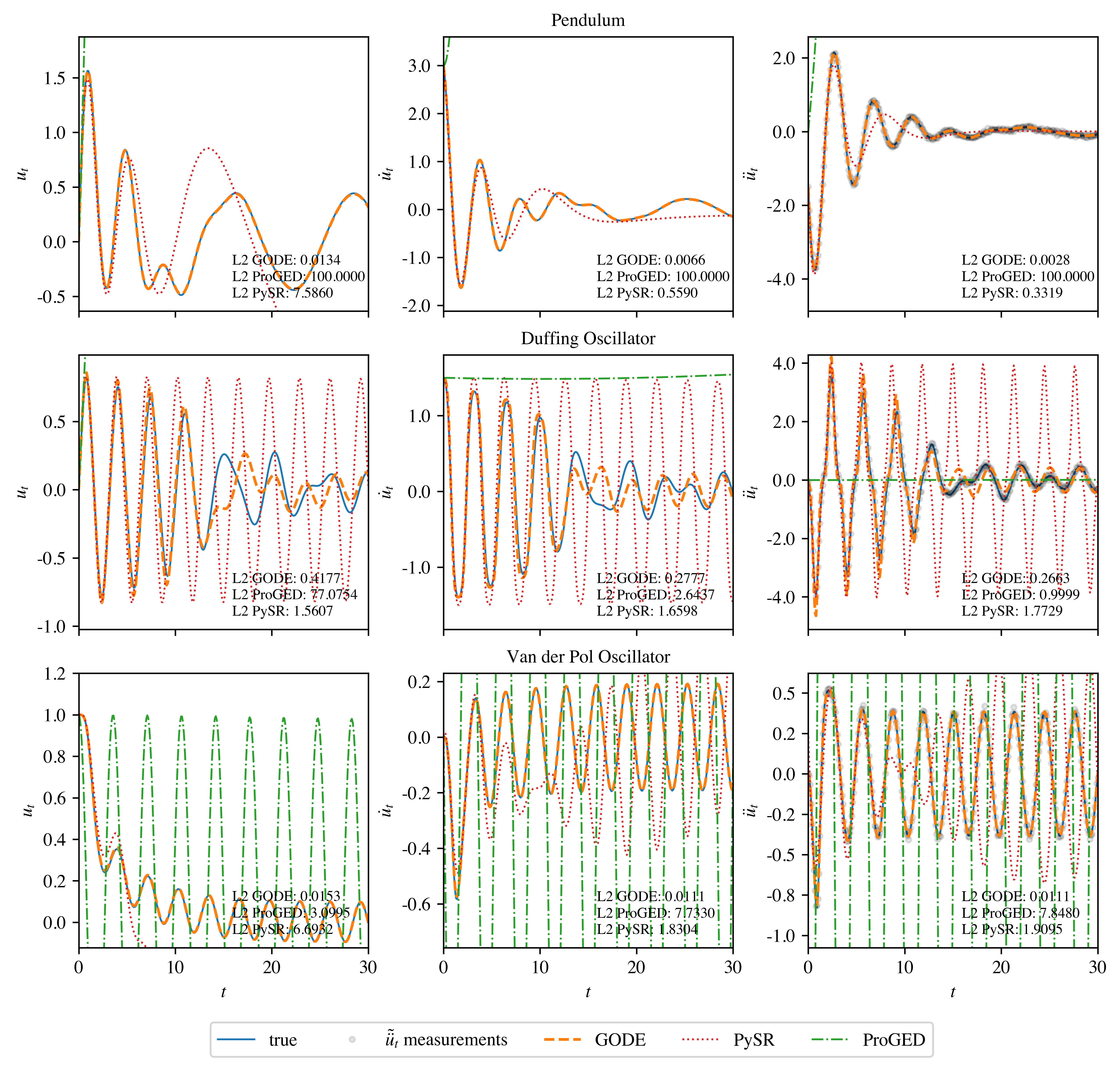}
    \caption{Solution trajectories of the true, measured, and predicted ODEs for the engineering examples. (The annotated L2 errors were capped at 100.)}
    \label{fig:b3_res}
\end{figure*}

\cref{fig:b3_res} illustrates the solution trajectories of both predicted and ground truth differential equations separately for $u_t$, $\dot{u}_t$, and $\ddot{u}_t$ for $t\in[0,30]$ with the L2 errors of each method, while \cref{tab:b3_odes} provides the initial conditions, sampling frequency $f_s$, time period, and symbolic mathematical expression. Although ProGED performed reasonably well on the simpler examples from the first benchmark (see \cref{ch3:subsubsec:b1_eval}), its performance on more complex inference tasks is far poorer. This might be partly attributed to its simple grammar template, yet it is still unexpected as the sought ODEs, such as the pendulum example, are not necessarily more complex than those in the second benchmark. Furthermore, the optimization of constants in ProGED uses differential evolution, which performs worse with an increasing number of constants, as the number of possible sets of constants explodes. In contrast, PySR encounters both difficulties in predicting simple and accurate expressions for these three inference tasks, highlighting the challenges of employing popular symbolic regression tools for engineering applications. Investigating the symbolic mathematical expressions more closely, shows that without additional prior specifications PySR does not avoid nesting of expressions, potentially hindering interpretability. In contrast, our GODE successfully infers the correct dynamics for two cases, the pendulum and the Duffing oscillator. However, in the latter case the constants are further away from the ground truth constants, explaining the errors in accuracy. As previously mentioned for ProGED, the optimization of constants becomes exponentially more challenging with increasing number of constants. When the term $\cos(0.00032t)$, which approximates $1$ for small $t$, is omitted, the dynamics of the Van der Pol oscillator can also be identified. For all examples, the predicted dynamics are fairly close to the true dynamical system, possibly benefiting from imposing the force term. An expert might further refine skeleton equations based on the best estimate by GODE to derive more accurate expressions, leveraging the outer optimization loop to infer the constants. Lastly, when approximating more complex dynamics, the performance of all three methods, PySR, ProGED, and GODE, depend on the initialization of the respective search algorithm. Although we present the best prediction out of five random initializations, there might be a more suitable initialization yielding more accurate results.

\begin{table*}[h!]
    \centering
    \caption{Three engineering examples with the predicted ODEs.}
    \renewcommand{\arraystretch}{1.5}
    \begin{tabular}{|p{3cm}|p{12.5cm}|}
    \hline
    Model & ODE \\
    \hline
    \hline
    Pendulum & $t\in[0,60],f_s=10,$ and $[u({t_0}), \dot{u}({t_0})]=[0.0,3.0]$\\
    True & $2\frac{d^2}{dt^2}u(t)+\frac{d}{dt}u(t)+5u(t)-2\sin(0.5t)=0$ \\
    ProGED & $\frac{d^2}{dt^2}u(t)-u(t)=0$ \\
    PySR best & $\frac{d^2}{dt^2}u(t)-\cos(0.77)^t\cdot\left(1.46-\frac{d}{dt}u(t)-(3.24+t)\cdot u(t)\right)=0$ \\
    GODE (ours) & $2.03\frac{d^2}{dt^2}u(t)+1.02\frac{d}{dt}u(t)+5.08u(t)-2\sin(0.5t)=0$\\
    \hline
    Duffing oscillator & $t\in[0,30],f_s=10,$ and $[u({t_0}), \dot{u}({t_0})]=[0.0,1.5]$\\
    True & $5\frac{d^2}{dt^2}u(t)+\frac{d}{dt}u(t)+7u(t)+25u^3(t)-\cos(2t)=0$ \\
    ProGED & $\frac{d^2}{dt^2}u(t)-0.00030t+0.0030=0$ \\
    PySR best & $\frac{d^2}{dt^2}u(t)+30.67u(t)-29.03\sin(u(t))=0$\\
    GODE (ours) & $2.50\frac{d^2}{dt^2}u(t)+0.78\frac{d}{dt}u(t)+0.44u(t)+17.88u^3(t)-\cos(2t)=0$ \\
    \hline
    Van der Pol oscillator & $t\in[0,30],f_s=10,$ and $[u({t_0}), \dot{u}({t_0})]=[1.0,0.0]$\\ 
    True & $\frac{d^2}{dt^2}u(t)+5\left(1-u^2(t)\right)\frac{d}{dt}u(t)+u(t)-\cos(2t)=0$ \\
    ProGED & $\frac{d^2}{dt^2}u(t)+3.14u(t)\cdot\cos(u^2(t))+u(t)+0.049=0$ \\
    PySR best & $\frac{d^2}{dt^2}u(t)-\sin\left((-0.71-u(t))\cdot\left(\sin(t-0.14)\cdot\left(0.56^t\frac{d}{dt}u(t)\right)+\cos(t-0.11)\right)\right)=0$\\
    GODE (ours) & $0.98\frac{d^2}{dt^2}u(t)+5.01\frac{d}{dt}u(t)-4.87\cos(0.00032t)\frac{d}{dt}u(t)\cdot u^2(t)+1.00u(t)-\cos(2t)=0$ \\
    \hline
    \end{tabular}
    \label{tab:b3_odes}
\end{table*}
\section{Discussion and conclusion}
\label{ch4:discussion}

\subsection{Summary}
The discovery of differential equations, as in the case of ODEs, is an ill-posed problem due to challenges such as noise, sparsity, and nonuniqueness. Symbolic regression methods range from discrete iterative algorithms, which construct symbolic expressions from primitives, to direct algorithms, which learn an operator between symbolic expressions and their numerical evaluations. In this work, we propose representing symbolic mathematical expressions using sequences of rules defined by formal grammars. These discrete sequences are embedded into a continuous latent space with the GVAE, which is explored using a stochastic iterative optimization algorithm, namely CMA-ES. We validate our approach by comparing GODE with three state-of-the-art methods: ODEFormer \cite{dascoli_odeformer_2023} (a transformer-based direct algorithm), PySR \cite{cranmer_interpretable_2023} (a genetic programming-based discrete search approach), and ProGED \cite{omejc_probabilistic_2024,brence_probabilistic_2021} (a grammar-based discrete search method) across various benchmarks. 

The first benchmark focused on first-order one-dimensional explicit ODEs, showing PySR as the most accurate method for simpler tasks, closely followed by ProGED and GODE. ODEFormer exhibited significantly lower sample and parameter efficiency than GODE, leading to its exclusion from further benchmarks. The second benchmark on first- and second-order linear or nonlinear ODEs demonstrated that our GODE discovered expressions that are both more parsimonious and accurate than those found by PySR and ProGED. The third benchmark investigated the discovery of nonlinear dynamics, commonly used in engineering applications, with only noisy acceleration measurements while the force term was known. In this scenario with partial information, GODE succeeded in both discovering accurate and parsimonious symbolic expressions. 

\subsection{Tokenization vs. grammar}
The first benchmark demonstrates that embedding discrete expressions using the ODEFormer algorithm is less sample- and parameter-efficient by orders of magnitude than GODE. As described in \cref{ch1:intro}, ODEFormer employs a tokenization strategy for input numerical data and decodes tokens to symbolic mathematical expressions \cite{dascoli_odeformer_2023}. Tokenization is a well-established technique in natural language processing and linguistics to convert text to tokens, typically represented as numerical vectors, and vice versa. However, a poorly chosen tokenization strategy leads to different types of ambiguities, where ambiguity here means that different sequences of tokens can form the same symbolic expression or the same input can lead to different outputs (stochastic ambiguity) \cite{gastaldi_foundations_2025}. This affects the consistency and reliability of the estimator and explains why ODEFormer requires so many samples to be accurate. It needs to resolve ambiguities through learning relations from data. Transformer-based models, such as the ODEFormer, require high computational power (approximately three days of training on one NVIDIA A100 GPU \cite{dascoli_odeformer_2023}) to provide accurate predictions. Moreover, an extensive dataset of 50 million trajectories needs to be considered, which means that to train the ODEFormer one employs a solver 50 million times, assuming that no expression is rejected, which is computationally prohibitive for complex systems. In contrast, our GODE allows us to introduce structure into the model through formal grammars. Formal grammars constitute an unambiguous representation of ODEs, and thus the deep learning algorithm does not require extracting relations from data, which dramatically increases the sample efficiency and decreases the model complexity. Although methods that include tokenizers have proven effective in large language models, due to both the higher complexity of natural languages, the abundance of text data, and the required flexibility of the model, these advantages do not necessarily apply to symbolic expressions. Stricter syntax rules and the adaptability to domain-specific problems can make rule-based approaches such as GODE more beneficial.

\subsection{Variance vs. bias and complexity vs. accuracy}
Symbolic regression aims not only to discover accurate mathematical expressions but also to ensure that they are interpretable \cite{schmidt_distilling_2009}. Algorithms, often based on genetic programming, such as PySR \cite{cranmer_interpretable_2023}, allow for any combination of primitives, resulting in high variance and potentially high complexity in possible candidate expressions. However, in fields such as science and engineering, researchers often possess prior knowledge which they wish to integrate into the symbolic regression process to guide the discovery. Such inductive biases can be introduced most easily in dictionary-based methods \cite{brunton_discovering_2016}, by defining a set of candidate expressions and using sparse regression to find the most parsimonious one. However, it is also the most restrictive, as with a growing number of candidate expressions the process becomes more complex, and SINDy only allows for linear combinations of candidate expressions. ProGED, which employs probabilistic CFGs, allows predefining probabilities of production rules to promote parsimony \cite{brence_probabilistic_2021}. Similarly, GODE introduces certain biases by restricting the generation of mathematical expressions through predefined rules, and it learns the probabilities of these rules. Various mechanisms, such as sparse regression with regularization techniques \cite{brunton_discovering_2016}, model selection criteria \cite{mangan_model_2017}, grammars \cite{todorovski_declarative_1997,brence_probabilistic_2021}, setting limits on the maximum length of generated expressions \cite{zembowicz_discovery_1992}, or manual examination of the discovered Pareto front \cite{schmidt_distilling_2009}, have been proposed to promote parsimony and reduce the complexity of mathematical expressions, which can hinder human interpretability. To balance the trade-off between complexity and accuracy, GODE leverages a multitude of these techniques, such as learned probabilistic grammars, the predefined maximum length of the sequence $N_\text{max}$, and the model selection criteria $\mathcal{L}_{IC}$ (\cref{eq:method_ic}).

\subsection{Outlook}
This section outlines potential future research opportunities for both GODE and the broader research area of symbolic regression and its application in scientific discovery. First, although GODE demonstrates greater sample and parameter efficiency than tokenizer-based models, it necessitates the creation of a grammar. The exploration of grammar induction (i.e., learning a grammar respectively a set of production rules) methods could alleviate this requirement. Researchers could provide a set of target mathematical expressions (e.g., various domain-specific ODEs), and the grammar induction technique could deduce a sparse grammar encapsulating these expressions. Moreover, the efficiency of GODE could be improved by including multiple modalities, such as the numerical solution trajectories and symbolic mathematical expressions, or by exploring alternative encoders or decoders. On the other hand, grammars can incorporate domain biases into the discovery process. Hence, a formal study investigating the types of constraints (e.g., soft vs. hard) that CFGs can enforce would be highly beneficial to our understanding. Although formal grammars offer an unambiguous representation of mathematical expressions, they do not directly address the issue of semantic ambiguity stemming from distributive, associative, and commutative properties of basic operations ($\{+,-,\cdot\}$) \cite{brence_probabilistic_2021}. Future research could explore methods and architectures, such as the use of attributes or embedding permutation invariance properties, to resolve these semantic ambiguities without solely relying on specific representations such as the canonical form. Lastly, the present study is limited to ODEs. Naturally, this could be extended to discovering partial differential equations (PDEs) and underlying conservation laws. One challenge in discovering PDEs is the higher complexity in solving them, often lacking analytical solutions and relying instead on numerical or deep learning-based methods, which in turn demands more sophisticated optimization objectives to be effective. Finally, such grammar-based methods could fundamentally transform downstream tasks in dynamical systems modeling, enabling more interpretable, robust, and generalizable tools for monitoring, control, and system identification.

\section*{Code availability}
The code will be made available online at the time of publication.

\section*{Acknowledgements}
This publication was made possible by an ETH AI Center doctoral fellowship to Karin L. Yu. Georgios Kissas would like to acknowledge support from Asuera Stiftung via the ETH Zurich Foundation and the ETH AI Center.

\bibliographystyle{IEEEtran}  
\bibliography{references}  

\begin{thebibliography}{10}
\providecommand{\url}[1]{#1}
\csname url@samestyle\endcsname
\providecommand{\newblock}{\relax}
\providecommand{\bibinfo}[2]{#2}
\providecommand{\BIBentrySTDinterwordspacing}{\spaceskip=0pt\relax}
\providecommand{\BIBentryALTinterwordstretchfactor}{4}
\providecommand{\BIBentryALTinterwordspacing}{\spaceskip=\fontdimen2\font plus
\BIBentryALTinterwordstretchfactor\fontdimen3\font minus \fontdimen4\font\relax}
\providecommand{\BIBforeignlanguage}[2]{{%
\expandafter\ifx\csname l@#1\endcsname\relax
\typeout{** WARNING: IEEEtran.bst: No hyphenation pattern has been}%
\typeout{** loaded for the language `#1'. Using the pattern for}%
\typeout{** the default language instead.}%
\else
\language=\csname l@#1\endcsname
\fi
#2}}
\providecommand{\BIBdecl}{\relax}
\BIBdecl

\bibitem{kuhn_structure_1997}
T.~S. Kuhn, \emph{The structure of scientific revolutions}.\hskip 1em plus 0.5em minus 0.4em\relax University of Chicago press Chicago, 1997, vol. 962.

\bibitem{kissas_language_2024}
\BIBentryALTinterwordspacing
G.~Kissas, S.~Mishra, E.~Chatzi, and L.~De~Lorenzis, ``\BIBforeignlanguage{en}{The language of hyperelastic materials},'' \emph{\BIBforeignlanguage{en}{Computer Methods in Applied Mechanics and Engineering}}, vol. 428, p. 117053, Aug. 2024. [Online]. Available: \url{https://linkinghub.elsevier.com/retrieve/pii/S0045782524003098}
\BIBentrySTDinterwordspacing

\bibitem{schmidt_distilling_2009}
\BIBentryALTinterwordspacing
M.~Schmidt and H.~Lipson, ``\BIBforeignlanguage{en}{Distilling {Free}-{Form} {Natural} {Laws} from {Experimental} {Data}},'' \emph{\BIBforeignlanguage{en}{Science}}, vol. 324, no. 5923, pp. 81--85, Apr. 2009. [Online]. Available: \url{https://www.science.org/doi/10.1126/science.1165893}
\BIBentrySTDinterwordspacing

\bibitem{popper_logic_1959}
K.~Popper, \emph{The logic of scientific discovery}.\hskip 1em plus 0.5em minus 0.4em\relax Routledge, 1959.

\bibitem{morgan_models_1999}
\BIBentryALTinterwordspacing
M.~S. Morgan and M.~Morrison, Eds., \emph{Models as {Mediators}: {Perspectives} on {Natural} and {Social} {Science}}, 1st~ed.\hskip 1em plus 0.5em minus 0.4em\relax Cambridge University Press, Oct. 1999. [Online]. Available: \url{https://www.cambridge.org/core/product/identifier/9780511660108/type/book}
\BIBentrySTDinterwordspacing

\bibitem{wang_scientific_2023}
\BIBentryALTinterwordspacing
H.~Wang, T.~Fu, Y.~Du, W.~Gao, K.~Huang, Z.~Liu, P.~Chandak, S.~Liu, P.~Van~Katwyk, A.~Deac, A.~Anandkumar, K.~Bergen, C.~P. Gomes, S.~Ho, P.~Kohli, J.~Lasenby, J.~Leskovec, T.-Y. Liu, A.~Manrai, D.~Marks, B.~Ramsundar, L.~Song, J.~Sun, J.~Tang, P.~Veličković, M.~Welling, L.~Zhang, C.~W. Coley, Y.~Bengio, and M.~Zitnik, ``\BIBforeignlanguage{en}{Scientific discovery in the age of artificial intelligence},'' \emph{\BIBforeignlanguage{en}{Nature}}, vol. 620, no. 7972, pp. 47--60, Aug. 2023. [Online]. Available: \url{https://www.nature.com/articles/s41586-023-06221-2}
\BIBentrySTDinterwordspacing

\bibitem{bongard_automated_2007}
\BIBentryALTinterwordspacing
J.~Bongard and H.~Lipson, ``\BIBforeignlanguage{en}{Automated reverse engineering of nonlinear dynamical systems},'' \emph{\BIBforeignlanguage{en}{Proceedings of the National Academy of Sciences}}, vol. 104, no.~24, pp. 9943--9948, Jun. 2007. [Online]. Available: \url{https://pnas.org/doi/full/10.1073/pnas.0609476104}
\BIBentrySTDinterwordspacing

\bibitem{brunton_discovering_2016}
\BIBentryALTinterwordspacing
S.~L. Brunton, J.~L. Proctor, and J.~N. Kutz, ``\BIBforeignlanguage{en}{Discovering governing equations from data by sparse identification of nonlinear dynamical systems},'' \emph{\BIBforeignlanguage{en}{Proceedings of the National Academy of Sciences}}, vol. 113, no.~15, pp. 3932--3937, Apr. 2016. [Online]. Available: \url{https://pnas.org/doi/full/10.1073/pnas.1517384113}
\BIBentrySTDinterwordspacing

\bibitem{dascoli_odeformer_2023}
\BIBentryALTinterwordspacing
S.~d'Ascoli, S.~Becker, A.~Mathis, P.~Schwaller, and N.~Kilbertus, ``{ODEFormer}: {Symbolic} {Regression} of {Dynamical} {Systems} with {Transformers},'' 2023, version Number: 1. [Online]. Available: \url{https://arxiv.org/abs/2310.05573}
\BIBentrySTDinterwordspacing

\bibitem{omejc_probabilistic_2024}
\BIBentryALTinterwordspacing
N.~Omejc, B.~Gec, J.~Brence, L.~Todorovski, and S.~Džeroski, ``\BIBforeignlanguage{en}{Probabilistic grammars for modeling dynamical systems from coarse, noisy, and partial data},'' \emph{\BIBforeignlanguage{en}{Machine Learning}}, vol. 113, no.~10, pp. 7689--7721, Oct. 2024. [Online]. Available: \url{https://link.springer.com/10.1007/s10994-024-06522-1}
\BIBentrySTDinterwordspacing

\bibitem{koza_genetic_1992}
J.~R. Koza, \emph{\BIBforeignlanguage{eng}{Genetic programming. 1: {On} the programming of computers by means of natural selection}}, ser. Complex adaptive systems.\hskip 1em plus 0.5em minus 0.4em\relax Cambridge, Mass.: MIT Press, 1992.

\bibitem{tsoulos_solving_2006}
\BIBentryALTinterwordspacing
I.~G. Tsoulos and I.~E. Lagaris, ``\BIBforeignlanguage{en}{Solving differential equations with genetic programming},'' \emph{\BIBforeignlanguage{en}{Genetic Programming and Evolvable Machines}}, vol.~7, no.~1, pp. 33--54, Mar. 2006. [Online]. Available: \url{http://link.springer.com/10.1007/s10710-006-7009-y}
\BIBentrySTDinterwordspacing

\bibitem{cranmer_discovering_2020}
\BIBentryALTinterwordspacing
M.~Cranmer, A.~Sanchez-Gonzalez, P.~Battaglia, R.~Xu, K.~Cranmer, D.~Spergel, and S.~Ho, ``Discovering {Symbolic} {Models} from {Deep} {Learning} with {Inductive} {Biases},'' 2020, version Number: 2. [Online]. Available: \url{https://arxiv.org/abs/2006.11287}
\BIBentrySTDinterwordspacing

\bibitem{cranmer_interpretable_2023}
\BIBentryALTinterwordspacing
M.~Cranmer, ``Interpretable {Machine} {Learning} for {Science} with {PySR} and {SymbolicRegression}.jl,'' 2023, version Number: 3. [Online]. Available: \url{https://arxiv.org/abs/2305.01582}
\BIBentrySTDinterwordspacing

\bibitem{mundhenk_symbolic_2021}
\BIBentryALTinterwordspacing
T.~N. Mundhenk, M.~Landajuela, R.~Glatt, C.~P. Santiago, D.~M. Faissol, and B.~K. Petersen, ``Symbolic {Regression} via {Neural}-{Guided} {Genetic} {Programming} {Population} {Seeding},'' 2021, version Number: 2. [Online]. Available: \url{https://arxiv.org/abs/2111.00053}
\BIBentrySTDinterwordspacing

\bibitem{petersen_deep_2019}
\BIBentryALTinterwordspacing
B.~K. Petersen, M.~Landajuela, T.~N. Mundhenk, C.~P. Santiago, S.~K. Kim, and J.~T. Kim, ``Deep symbolic regression: {Recovering} mathematical expressions from data via risk-seeking policy gradients,'' 2019, publisher: arXiv Version Number: 4. [Online]. Available: \url{https://arxiv.org/abs/1912.04871}
\BIBentrySTDinterwordspacing

\bibitem{todorovski_declarative_1997}
L.~Todorovski and S.~Džeroski, ``Declarative bias in equation discovery,'' in \emph{Proceedings of the 14th {International} {Coference} on {Machine} {Learning}. {ICML}.}, 1997, pp. 376--384.

\bibitem{brence_probabilistic_2021}
\BIBentryALTinterwordspacing
J.~Brence, L.~Todorovski, and S.~Džeroski, ``\BIBforeignlanguage{en}{Probabilistic grammars for equation discovery},'' \emph{\BIBforeignlanguage{en}{Knowledge-Based Systems}}, vol. 224, p. 107077, Jul. 2021. [Online]. Available: \url{https://linkinghub.elsevier.com/retrieve/pii/S0950705121003403}
\BIBentrySTDinterwordspacing

\bibitem{messenger_weak_2021}
\BIBentryALTinterwordspacing
D.~A. Messenger and D.~M. Bortz, ``\BIBforeignlanguage{en}{Weak {SINDy}: {Galerkin}-{Based} {Data}-{Driven} {Model} {Selection}},'' \emph{\BIBforeignlanguage{en}{Multiscale Modeling \& Simulation}}, vol.~19, no.~3, pp. 1474--1497, Jan. 2021. [Online]. Available: \url{https://epubs.siam.org/doi/10.1137/20M1343166}
\BIBentrySTDinterwordspacing

\bibitem{lample_deep_2019}
\BIBentryALTinterwordspacing
G.~Lample and F.~Charton, ``Deep {Learning} for {Symbolic} {Mathematics},'' 2019, version Number: 1. [Online]. Available: \url{https://arxiv.org/abs/1912.01412}
\BIBentrySTDinterwordspacing

\bibitem{kamienny_end--end_2022}
\BIBentryALTinterwordspacing
P.-A. Kamienny, S.~d'Ascoli, G.~Lample, and F.~Charton, ``End-to-end symbolic regression with transformers,'' 2022, version Number: 1. [Online]. Available: \url{https://arxiv.org/abs/2204.10532}
\BIBentrySTDinterwordspacing

\bibitem{kamienny_deep_2023}
\BIBentryALTinterwordspacing
P.-A. Kamienny, G.~Lample, S.~Lamprier, and M.~Virgolin, ``Deep {Generative} {Symbolic} {Regression} with {Monte}-{Carlo}-{Tree}-{Search},'' in \emph{Proceedings of the 40th {International} {Conference} on {Machine} {Learning}}, ser. Proceedings of {Machine} {Learning} {Research}, A.~Krause, E.~Brunskill, K.~Cho, B.~Engelhardt, S.~Sabato, and J.~Scarlett, Eds., vol. 202.\hskip 1em plus 0.5em minus 0.4em\relax PMLR, Jul. 2023, pp. 15\,655--15\,668. [Online]. Available: \url{https://proceedings.mlr.press/v202/kamienny23a.html}
\BIBentrySTDinterwordspacing

\bibitem{becker_predicting_2023}
\BIBentryALTinterwordspacing
S.~Becker, M.~Klein, A.~Neitz, G.~Parascandolo, and N.~Kilbertus, ``Predicting {Ordinary} {Differential} {Equations} with {Transformers},'' in \emph{Proceedings of the 40th {International} {Conference} on {Machine} {Learning}}, ser. Proceedings of {Machine} {Learning} {Research}, A.~Krause, E.~Brunskill, K.~Cho, B.~Engelhardt, S.~Sabato, and J.~Scarlett, Eds., vol. 202.\hskip 1em plus 0.5em minus 0.4em\relax PMLR, Jul. 2023, pp. 1978--2002. [Online]. Available: \url{https://proceedings.mlr.press/v202/becker23a.html}
\BIBentrySTDinterwordspacing

\bibitem{dascoli_deep_2022}
\BIBentryALTinterwordspacing
S.~d'Ascoli, P.-A. Kamienny, G.~Lample, and F.~Charton, ``Deep {Symbolic} {Regression} for {Recurrent} {Sequences},'' 2022, version Number: 2. [Online]. Available: \url{https://arxiv.org/abs/2201.04600}
\BIBentrySTDinterwordspacing

\bibitem{sennrich_neural_2015}
\BIBentryALTinterwordspacing
R.~Sennrich, B.~Haddow, and A.~Birch, ``Neural {Machine} {Translation} of {Rare} {Words} with {Subword} {Units},'' 2015, version Number: 5. [Online]. Available: \url{https://arxiv.org/abs/1508.07909}
\BIBentrySTDinterwordspacing

\bibitem{hopcroft_introduction_1999}
J.~E. Hopcroft and J.~D. Ullman, \emph{\BIBforeignlanguage{eng}{Introduction to automata theory, languages, and computation}}, 39th~ed., ser. Addison-{Wesley} series in computer science.\hskip 1em plus 0.5em minus 0.4em\relax Reading, Mass.: Addison-Wesley, 1999.

\bibitem{kusner_grammar_2017}
\BIBentryALTinterwordspacing
M.~J. Kusner, B.~Paige, and J.~M. Hernández-Lobato, ``Grammar {Variational} {Autoencoder},'' in \emph{Proceedings of the 34th {International} {Conference} on {Machine} {Learning}}, ser. Proceedings of {Machine} {Learning} {Research}, D.~Precup and Y.~W. Teh, Eds., vol.~70.\hskip 1em plus 0.5em minus 0.4em\relax PMLR, Aug. 2017, pp. 1945--1954. [Online]. Available: \url{https://proceedings.mlr.press/v70/kusner17a.html}
\BIBentrySTDinterwordspacing

\bibitem{oikonomou_neuro-symbolic_2025}
\BIBentryALTinterwordspacing
O.~Oikonomou, L.~Lingsch, D.~Grund, S.~Mishra, and G.~Kissas, ``Neuro-{Symbolic} {AI} for {Analytical} {Solutions} of {Differential} {Equations},'' 2025, version Number: 1. [Online]. Available: \url{https://arxiv.org/abs/2502.01476}
\BIBentrySTDinterwordspacing

\bibitem{bird_natural_2009}
\BIBentryALTinterwordspacing
S.~Bird, E.~Klein, and E.~Loper, \emph{Natural {Language} {Processing} with {Python}: {Analyzing} {Text} with the {Natural} {Language} {Toolkit}}.\hskip 1em plus 0.5em minus 0.4em\relax O'Reilly Media, Inc., Jun. 2009. [Online]. Available: \url{https://www.nltk.org/book/}
\BIBentrySTDinterwordspacing

\bibitem{kingma_auto-encoding_2014}
\BIBentryALTinterwordspacing
D.~P. Kingma and M.~Welling, ``Auto-{Encoding} {Variational} {Bayes},'' in \emph{Proceedings of the {International} {Conference} on {Learning} {Representations} ({ICLR})}.\hskip 1em plus 0.5em minus 0.4em\relax arXiv, 2014, version Number: 11. [Online]. Available: \url{https://arxiv.org/abs/1312.6114}
\BIBentrySTDinterwordspacing

\bibitem{cho_learning_2014}
\BIBentryALTinterwordspacing
K.~Cho, B.~Van~Merrienboer, C.~Gulcehre, D.~Bahdanau, F.~Bougares, H.~Schwenk, and Y.~Bengio, ``\BIBforeignlanguage{en}{Learning {Phrase} {Representations} using {RNN} {Encoder}–{Decoder} for {Statistical} {Machine} {Translation}},'' in \emph{\BIBforeignlanguage{en}{Proceedings of the 2014 {Conference} on {Empirical} {Methods} in {Natural} {Language} {Processing} ({EMNLP})}}.\hskip 1em plus 0.5em minus 0.4em\relax Doha, Qatar: Association for Computational Linguistics, 2014, pp. 1724--1734. [Online]. Available: \url{http://aclweb.org/anthology/D14-1179}
\BIBentrySTDinterwordspacing

\bibitem{clevert_fast_2015}
\BIBentryALTinterwordspacing
D.-A. Clevert, T.~Unterthiner, and S.~Hochreiter, ``Fast and {Accurate} {Deep} {Network} {Learning} by {Exponential} {Linear} {Units} ({ELUs}),'' 2015, version Number: 5. [Online]. Available: \url{https://arxiv.org/abs/1511.07289}
\BIBentrySTDinterwordspacing

\bibitem{kingma_adam_2014}
\BIBentryALTinterwordspacing
D.~P. Kingma and J.~Ba, ``Adam: {A} {Method} for {Stochastic} {Optimization},'' 2014, version Number: 9. [Online]. Available: \url{https://arxiv.org/abs/1412.6980}
\BIBentrySTDinterwordspacing

\bibitem{hansen_cma_2016}
\BIBentryALTinterwordspacing
N.~Hansen, ``The {CMA} {Evolution} {Strategy}: {A} {Tutorial},'' 2016, version Number: 2. [Online]. Available: \url{https://arxiv.org/abs/1604.00772}
\BIBentrySTDinterwordspacing

\bibitem{hansen_completely_2001}
\BIBentryALTinterwordspacing
N.~Hansen and A.~Ostermeier, ``\BIBforeignlanguage{en}{Completely {Derandomized} {Self}-{Adaptation} in {Evolution} {Strategies}},'' \emph{\BIBforeignlanguage{en}{Evolutionary Computation}}, vol.~9, no.~2, pp. 159--195, Jun. 2001. [Online]. Available: \url{https://direct.mit.edu/evco/article/9/2/159-195/892}
\BIBentrySTDinterwordspacing

\bibitem{hansen_reducing_2003}
\BIBentryALTinterwordspacing
N.~Hansen, S.~D. Müller, and P.~Koumoutsakos, ``\BIBforeignlanguage{en}{Reducing the {Time} {Complexity} of the {Derandomized} {Evolution} {Strategy} with {Covariance} {Matrix} {Adaptation} ({CMA}-{ES})},'' \emph{\BIBforeignlanguage{en}{Evolutionary Computation}}, vol.~11, no.~1, pp. 1--18, Mar. 2003. [Online]. Available: \url{https://direct.mit.edu/evco/article/11/1/1-18/1139}
\BIBentrySTDinterwordspacing

\bibitem{hansen_cma-espycma_2024}
\BIBentryALTinterwordspacing
N.~Hansen, Y.~Akimoto, and P.~Baudis, ``{CMA}-{ES}/pycma: r4.0.0,'' Sep. 2024. [Online]. Available: \url{https://zenodo.org/doi/10.5281/zenodo.2559634}
\BIBentrySTDinterwordspacing

\bibitem{nelder_simplex_1965}
\BIBentryALTinterwordspacing
J.~A. Nelder and R.~Mead, ``\BIBforeignlanguage{en}{A {Simplex} {Method} for {Function} {Minimization}},'' \emph{\BIBforeignlanguage{en}{The Computer Journal}}, vol.~7, no.~4, pp. 308--313, Jan. 1965. [Online]. Available: \url{https://academic.oup.com/comjnl/article-lookup/doi/10.1093/comjnl/7.4.308}
\BIBentrySTDinterwordspacing

\bibitem{virtanen_scipy_2020}
\BIBentryALTinterwordspacing
P.~Virtanen, R.~Gommers, T.~E. Oliphant, M.~Haberland, T.~Reddy, D.~Cournapeau, E.~Burovski, P.~Peterson, W.~Weckesser, J.~Bright, S.~J. Van Der~Walt, M.~Brett, J.~Wilson, K.~J. Millman, N.~Mayorov, A.~R.~J. Nelson, E.~Jones, R.~Kern, E.~Larson, C.~J. Carey, I.~Polat, Y.~Feng, E.~W. Moore, J.~VanderPlas, D.~Laxalde, J.~Perktold, R.~Cimrman, I.~Henriksen, E.~A. Quintero, C.~R. Harris, A.~M. Archibald, A.~H. Ribeiro, F.~Pedregosa, P.~Van~Mulbregt, {SciPy 1.0 Contributors}, A.~Vijaykumar, A.~P. Bardelli, A.~Rothberg, A.~Hilboll, A.~Kloeckner, A.~Scopatz, A.~Lee, A.~Rokem, C.~N. Woods, C.~Fulton, C.~Masson, C.~Häggström, C.~Fitzgerald, D.~A. Nicholson, D.~R. Hagen, D.~V. Pasechnik, E.~Olivetti, E.~Martin, E.~Wieser, F.~Silva, F.~Lenders, F.~Wilhelm, G.~Young, G.~A. Price, G.-L. Ingold, G.~E. Allen, G.~R. Lee, H.~Audren, I.~Probst, J.~P. Dietrich, J.~Silterra, J.~T. Webber, J.~Slavič, J.~Nothman, J.~Buchner, J.~Kulick, J.~L. Schönberger, J.~V. De~Miranda~Cardoso, J.~Reimer, J.~Harrington, J.~L.~C. Rodríguez,
  J.~Nunez-Iglesias, J.~Kuczynski, K.~Tritz, M.~Thoma, M.~Newville, M.~Kümmerer, M.~Bolingbroke, M.~Tartre, M.~Pak, N.~J. Smith, N.~Nowaczyk, N.~Shebanov, O.~Pavlyk, P.~A. Brodtkorb, P.~Lee, R.~T. McGibbon, R.~Feldbauer, S.~Lewis, S.~Tygier, S.~Sievert, S.~Vigna, S.~Peterson, S.~More, T.~Pudlik, T.~Oshima, T.~J. Pingel, T.~P. Robitaille, T.~Spura, T.~R. Jones, T.~Cera, T.~Leslie, T.~Zito, T.~Krauss, U.~Upadhyay, Y.~O. Halchenko, and Y.~Vázquez-Baeza, ``\BIBforeignlanguage{en}{{SciPy} 1.0: fundamental algorithms for scientific computing in {Python}},'' \emph{\BIBforeignlanguage{en}{Nature Methods}}, vol.~17, no.~3, pp. 261--272, Mar. 2020. [Online]. Available: \url{https://www.nature.com/articles/s41592-019-0686-2}
\BIBentrySTDinterwordspacing

\bibitem{elfwing_sigmoid-weighted_2017}
\BIBentryALTinterwordspacing
S.~Elfwing, E.~Uchibe, and K.~Doya, ``Sigmoid-{Weighted} {Linear} {Units} for {Neural} {Network} {Function} {Approximation} in {Reinforcement} {Learning},'' 2017, version Number: 3. [Online]. Available: \url{https://arxiv.org/abs/1702.03118}
\BIBentrySTDinterwordspacing

\bibitem{paszke_automatic_2017}
A.~Paszke, S.~Gross, S.~Chintala, G.~Chanan, E.~Yang, Z.~DeVito, Z.~Lin, A.~Desmaison, L.~Antiga, and A.~Lerer, ``Automatic differentiation in pytorch,'' 2017.

\bibitem{symengine_development_team_symengine_2023}
\BIBentryALTinterwordspacing
{SymEngine Development Team}, ``{SymEngine} {Python} {Wrappers},'' Dec. 2023, version 0.11.0. [Online]. Available: \url{https://github.com/symengine/symengine.py}
\BIBentrySTDinterwordspacing

\bibitem{meurer_sympy_2017}
\BIBentryALTinterwordspacing
A.~Meurer, C.~P. Smith, M.~Paprocki, O.~Čertík, S.~B. Kirpichev, M.~Rocklin, A.~Kumar, S.~Ivanov, J.~K. Moore, S.~Singh, T.~Rathnayake, S.~Vig, B.~E. Granger, R.~P. Muller, F.~Bonazzi, H.~Gupta, S.~Vats, F.~Johansson, F.~Pedregosa, M.~J. Curry, A.~R. Terrel, S.~Roučka, A.~Saboo, I.~Fernando, S.~Kulal, R.~Cimrman, and A.~Scopatz, ``\BIBforeignlanguage{en}{{SymPy}: symbolic computing in {Python}},'' \emph{\BIBforeignlanguage{en}{PeerJ Computer Science}}, vol.~3, p. e103, Jan. 2017. [Online]. Available: \url{https://peerj.com/articles/cs-103}
\BIBentrySTDinterwordspacing

\bibitem{hyeonjung_survey_2023}
\BIBentryALTinterwordspacing
Hyeonjung, Jung, J.~Gupta, B.~Jayaprakash, M.~Eagon, H.~P. Selvam, C.~Molnar, W.~Northrop, and S.~Shekhar, ``A {Survey} on {Solving} and {Discovering} {Differential} {Equations} {Using} {Deep} {Neural} {Networks},'' 2023, version Number: 2. [Online]. Available: \url{https://arxiv.org/abs/2304.13807}
\BIBentrySTDinterwordspacing

\bibitem{scholl_symbolic_2022}
\BIBentryALTinterwordspacing
P.~Scholl, A.~Bacho, H.~Boche, and G.~Kutyniok, ``Symbolic {Recovery} of {Differential} {Equations}: {The} {Identifiability} {Problem},'' 2022, version Number: 9. [Online]. Available: \url{https://arxiv.org/abs/2210.08342}
\BIBentrySTDinterwordspacing

\bibitem{riolo_symbolic_2010}
\BIBentryALTinterwordspacing
M.~Schmidt and H.~Lipson, ``\BIBforeignlanguage{en}{Symbolic {Regression} of {Implicit} {Equations}},'' in \emph{\BIBforeignlanguage{en}{Genetic {Programming} {Theory} and {Practice} {VII}}}, R.~Riolo, U.-M. O'Reilly, and T.~McConaghy, Eds.\hskip 1em plus 0.5em minus 0.4em\relax Boston, MA: Springer US, 2010, pp. 73--85, series Title: Genetic and Evolutionary Computation. [Online]. Available: \url{https://link.springer.com/10.1007/978-1-4419-1626-6_5}
\BIBentrySTDinterwordspacing

\bibitem{ansel_pytorch_2024}
\BIBentryALTinterwordspacing
J.~Ansel, E.~Yang, H.~He, N.~Gimelshein, A.~Jain, M.~Voznesensky, B.~Bao, P.~Bell, D.~Berard, E.~Burovski, G.~Chauhan, A.~Chourdia, W.~Constable, A.~Desmaison, Z.~DeVito, E.~Ellison, W.~Feng, J.~Gong, M.~Gschwind, B.~Hirsh, S.~Huang, K.~Kalambarkar, L.~Kirsch, M.~Lazos, M.~Lezcano, Y.~Liang, J.~Liang, Y.~Lu, C.~K. Luk, B.~Maher, Y.~Pan, C.~Puhrsch, M.~Reso, M.~Saroufim, M.~Y. Siraichi, H.~Suk, S.~Zhang, M.~Suo, P.~Tillet, X.~Zhao, E.~Wang, K.~Zhou, R.~Zou, X.~Wang, A.~Mathews, W.~Wen, G.~Chanan, P.~Wu, and S.~Chintala, ``\BIBforeignlanguage{en}{{PyTorch} 2: {Faster} {Machine} {Learning} {Through} {Dynamic} {Python} {Bytecode} {Transformation} and {Graph} {Compilation}},'' in \emph{\BIBforeignlanguage{en}{Proceedings of the 29th {ACM} {International} {Conference} on {Architectural} {Support} for {Programming} {Languages} and {Operating} {Systems}, {Volume} 2}}.\hskip 1em plus 0.5em minus 0.4em\relax La Jolla CA USA: ACM, Apr. 2024, pp. 929--947. [Online]. Available:
  \url{https://dl.acm.org/doi/10.1145/3620665.3640366}
\BIBentrySTDinterwordspacing

\bibitem{duffing_ingenieur_1918}
G.~Duffing, ``\BIBforeignlanguage{de}{Ingenieur: {Erzwungene} {Schwingungen} bei veränderlicher {Eigenfrequenz} und ihre technische {Bedeutung}},'' \emph{\BIBforeignlanguage{de}{Sammlung Vieweg}}, no. Heft 41/42, pp. vi+334, 1918.

\bibitem{van_der_pol_lxxxviii_1926}
\BIBentryALTinterwordspacing
B.~Van Der~Pol, ``\BIBforeignlanguage{en}{{LXXXVIII}. \textit{{On} “relaxation-oscillations”}},'' \emph{\BIBforeignlanguage{en}{The London, Edinburgh, and Dublin Philosophical Magazine and Journal of Science}}, vol.~2, no.~11, pp. 978--992, Nov. 1926. [Online]. Available: \url{http://www.tandfonline.com/doi/abs/10.1080/14786442608564127}
\BIBentrySTDinterwordspacing

\bibitem{gastaldi_foundations_2025}
\BIBentryALTinterwordspacing
J.~L. Gastaldi, J.~Terilla, L.~Malagutti, B.~DuSell, T.~Vieira, and R.~Cotterell, ``The {Foundations} of {Tokenization}: {Statistical} and {Computational} {Concerns},'' in \emph{Proceedings of the {International} {Conference} on {Learning} {Representations} ({ICLR})}.\hskip 1em plus 0.5em minus 0.4em\relax arXiv, 2025, version Number: 3. [Online]. Available: \url{https://arxiv.org/abs/2407.11606}
\BIBentrySTDinterwordspacing

\bibitem{mangan_model_2017}
\BIBentryALTinterwordspacing
N.~M. Mangan, J.~N. Kutz, S.~L. Brunton, and J.~L. Proctor, ``\BIBforeignlanguage{en}{Model selection for dynamical systems via sparse regression and information criteria},'' \emph{\BIBforeignlanguage{en}{Proceedings of the Royal Society A: Mathematical, Physical and Engineering Sciences}}, vol. 473, no. 2204, p. 20170009, Aug. 2017. [Online]. Available: \url{https://royalsocietypublishing.org/doi/10.1098/rspa.2017.0009}
\BIBentrySTDinterwordspacing

\bibitem{zembowicz_discovery_1992}
R.~Zembowicz and J.~M. Zytkow, ``Discovery of equations: experimental evaluation of convergence,'' in \emph{Proceedings of the {Tenth} {National} {Conference} on {Artificial} {Intelligence}}, ser. {AAAI}'92.\hskip 1em plus 0.5em minus 0.4em\relax AAAI Press, 1992, pp. 70--75, place: San Jose, California.

\end{thebibliography}

\clearpage
\appendix
\section*{Appendix}
\section{Problem formulation for each method}
\label{app:problem_form_method}

This section summarizes what kind of problem each method attempts to solve, what kind of input it expects, and what kind of output it provides. 

\textbf{ODEFormer} \cite{dascoli_odeformer_2023} solves explicit ODEs:
\begin{align*}
    \dot{u}(t) & = f(u(t)) \\
    \text{INPUT: }\quad & t\text{, }u_t \\
    \text{OUTPUT: }\quad & f(u(t)) \rightarrow \text{ODE: } \dot{u}(t)-f(u(t))
\end{align*}

\textbf{PySR} \cite{cranmer_interpretable_2023} solves function approximation:
\begin{align*}
    y & = f(X) \\
    \text{INPUT: }\quad & \text{matrix $X$, in our case $X=[t, u_t], y = \dot{u}_t$ for first-order ODEs,} \\
    & \text{and $X=[t, u_t, \dot{u}_t], y = \ddot{u}_t$ for second-order ODEs} \\
    \text{OUTPUT: }\quad & f(X) \rightarrow \text{ODE: } y-f(X)
\end{align*}

\textbf{ProGED} \cite{brence_probabilistic_2021, omejc_probabilistic_2024} solves function approximation:
\begin{align*}
    Y & = f(X) \\
    \text{INPUT: }\quad & \text{Pandas dataframe with $[t, u_t, \dot{u}_t]$ for first-order,} \\ &
    \text{and $[t, u_t, \dot{u}_t, \ddot{u}_t]$ for second-order ODEs} \\
    & \text{Definition of the left- and right-hand side} \\
    \text{OUTPUT: }\quad & f(X) \rightarrow \text{ODE: } y-f(X)
\end{align*}

\textbf{GODE} (ours) solves for the first benchmark:
\begin{align*}
    \dot{u}(t) & = f(u(t)) \\
    \text{INPUT: } \quad & [t, u_t, \dot{u}_t] \\
    \text{OUTPUT: }\quad &f(u(t)) \rightarrow \text{ODE: } \dot{u}(t)-f(u(t))
\end{align*}

\textbf{GODE} (ours) solves for the other examples :
\begin{align*}
    \mathcal{D}(u(t)) - F(t) & = 0 \\
    \text{INPUT: } \quad & [t, u_t, \dot{u}_t] \text{ for first-order ODEs, and order = 1,} \\ 
    & [t, u_t, \dot{u}_t, \ddot{u}_t] \text{ for second-order ODEs, and order = 2,} \\
    & \text{if force term is available, additionally }F(t)\\
    \text{OUTPUT: }\quad & \text{directly the symbolic equation: } \mathcal{D}(u(t)) - F(t) = 0
\end{align*}

\section{Details on the benchmark of one-dimensional explicit ODEs}
\label{app:Bench1}
This section offers some further details on the dataset generation and experiments of the first benchmark on one-dimensional explicit ODEs of \cref{ch3:subsec:bench1}.

The underlying grammar for both the training dataset generation and GVAE is listed in \cref{tab:app1:cfg_bench1}. The last rule is the padding rule described in \cref{ch2:subsec:library}. \cref{tab:app1:odebench,tab:app1:add} list the 30 tested examples with the domain for time $t$, the initial value at $t_0$, and the sampling frequency. 

\begin{table*}[h!]
    \caption{Context-free grammar for the first benchmark.}
    \label{tab:app1:cfg_bench1}
    \lstset{basicstyle=\small}
    \centering
    \begin{lstlisting}[breaklines]
        start -> term '*' expr '+' start | term '*' expr
        expr -> '1' | '1' mop '(' small_expr ')' | F '^' int | F '^' int mop '(' small_expr ')' | func | F '^' int mop func | func mop func
        func -> 'sin' '(' inner_func ')' | 'cos' '(' inner_func ')' | 'exp' '(' inner_func ')' | 'log' '(' inner_func ')'
        inner_func -> F mop term
        small_expr -> term '+' F '^' int mop term
        mop -> '*' | '/'
        term ->  num | num '*' var
        F -> 'u'
        var -> 't'
        num -> 'C'
        int -> '1' | '2' | '3' | '4' | '5'
        Nothing -> None
    \end{lstlisting}
\end{table*}


\begin{table*}[t!]
    \renewcommand{\arraystretch}{1.75}
    \centering
    \caption{Test examples from ODEBench \citep{dascoli_odeformer_2023} used in the first benchmark.}
    \begin{tabular}{|l|l|l|l|l|}
    \hline
    Name & Ordinary differential equation & Time domain & Init. values & Samp. freq. \\
    \hline
    \hline
    ID1 & $\frac{du(t)}{dt}+\frac{1}{1.2\cdot 2.31}u(t)=\frac{0.7}{2.31}$ & [$0.1$, $15.0$] & $10.0$ & $8$\\
    ID2 & $\frac{du(t)}{dt}-0.23u(t)=0$ & [$4.0$, $17.3$] & $4.78$ & $5$\\
    ID3 & $\frac{du(t)}{dt}-0.79u(t)+\frac{0.79}{74.3}u(t)^2=0$ & [$0.5$, $25.0$] & $7.3$ & $5$\\
    ID4 & $\frac{du(t)}{dt}-\frac{1}{1+\exp{\left(0.5-\frac{u(t)}{0.96}\right)}}=-0.5$  & [$1.5$, $9.0$] & $0.8$ & $9$\\
    ID5 & $\frac{du(t)}{dt}+0.0021175u(t)^2=9.81$ & [$2.0$, $23.0$] & $0.5$ & $6$\\
    ID6 & $\frac{du(t)}{dt}-2.1u(t)+0.5u(t)^2=0$ & [$0.3$, $5.5$] & $0.13$ & $10$\\
    ID7 & $\frac{du(t)}{dt}-0.032u(t)\cdot\log(2.29u(t))=0$ & [$1.5$, $12.9$] & $1.73$ & $10$\\
    ID8 & $\frac{du(t)}{dt}-0.14u(t)\cdot\left(\frac{u(t)}{4.4}-1\right)\cdot\left(1-\frac{u(t)}{130.0}\right)=0$ & [$1.8$, $9.1$] & $6.123$ & $15$\\
    ID9 & $\frac{du(t)}{dt}+0.60u(t)=0.32$ & [$0.3$, $24.1$] & $0.14$ & $8$\\
    ID10 & $\frac{du(t)}{dt}-0.2u(t)^{1.2}\cdot(1-u(t))+0.8u(t)(1-u(t))^{1.2}=0$ & [$1.1$, $14.3$] & $0.83$ & $10$\\
    ID11 & $\frac{du(t)}{dt}+u(t)^3=0$ & [$2.0$, $7.0$] & $3.4$ & $20$\\
    ID12 & $\frac{du(t)}{dt}-1.8u(t)+0.1107u(t)^2=0$ & [$4.3$, $18.4$] & $11.0$ & $8$\\
    ID13 & $\frac{du(t)}{dt}-0.0981\cdot\left(9.7\cos(u(t))-1\right)\cdot\sin(u(t))=0$ & [$0.1$, $18.9$] & $2.4$ & $10$\\
    ID14 & $\frac{du(t)}{dt}-0.78u(t)\cdot\left(1-\frac{u(t)}{81.0}\right)+0.9\frac{u(t)^2}{21.2^2+u(t)^2}=0$ & [$1.4$, $8.2$] & $2.76$ & $13$\\
    ID15 & $\frac{du(t)}{dt}-0.4u(t)\cdot\left(1-\frac{u(t)}{95.0}\right)+\frac{u(t)^2}{1+u(t)^2}=0$ & [$3.8$, $13.8$] & $44.3$ & $9$\\
    ID16 & $\frac{du(t)}{dt}-0.1u(t)-0.04u(t)^3+0.001u(t)^5=0$ & [$0.8$, $9.2$] & $0.94$ & $18$\\
    ID17 & $\frac{du(t)}{dt}-0.4u(t)\cdot\left(1-\frac{u(t)}{100.0}\right)=-0.3$ & [$1.4$, $15.2$] & $14.3$ & $9$\\
    ID18 & $\frac{du(t)}{dt}-0.4u(t)\cdot\left(1-\frac{u(t)}{100.0}\right)+0.24\frac{u(t)}{50.0+u(t)}=0$ & [$1.8$, $17.0$] & $21.1$ & $6$\\
    ID19 & $\frac{du(t)}{dt}+0.08u(t)\frac{u(t)}{0.8+u(t)}+u(t)\left(1-u(t)\right)\cdot=0$ & [$2.5$, $8.0$] & $0.13$ & $20$\\
    ID20 & $\frac{du(t)}{dt}+0.55u(t)+\frac{u(t)^2}{u(t)^2+1.0}=0.1$ & [$1.5$, $11.3$] & $0.002$ & $16$\\
    ID21 & $\frac{du(t)}{dt}-0.2u(t)+\exp{(u(t))}=1.2$ & [$2.0$, $9.6$] & $0.0$ & $18$\\
    ID22 & $\frac{du(t)}{dt}-0.4\frac{u(t)^5}{123.0+u(t)^5}+0.89u(t)=1.4$ & [$2.8$, $7.9$] & $3.1$ & $15$\\
    ID23 & $\frac{du(t)}{dt}+\sin(u(t))=0.21$ & [$4.9$, $33.8$] & $-2.74$ & $10$\\
    \hline
    \end{tabular}
    \label{tab:app1:odebench}
\end{table*}

\begin{table*}[h!]
    \renewcommand{\arraystretch}{1.75}
    \centering
    \caption{Additional tested examples of the first benchmark.}
    \begin{tabular}{|l|l|l|l|l|}
    \hline
    Name & Ordinary differential equation & Time domain & Init. values & Samp. freq. \\
    \hline
    \hline
    ID24 & $\frac{du(t)}{dt}+\sin{\left(0.2t\cdot u(t)\right)}=0.21$ & [$2.9$, $13.8$] & $-1.89$ & $12$\\
    ID25 & $\frac{du(t)}{dt}+u(t)=-0.9832\cos{\left(0.132t\right)}$ & [$0.9$, $52.8$] & $1.89$ & $4$\\
    ID26 & $\frac{du(t)}{dt}-\exp{\left(\frac{1.73}{u(t)}\right)}=-7.928$ & [$1.5$, $13.8$] & $2.92$ & $12$\\
    ID27 & $\frac{du(t)}{dt}-\exp{\left(1.03u(t)\right)}=\cos(0.2t)-7.928$ & [$0.8$, $12.2$] & $0.05$ & $12$\\
    ID28 & $\frac{du(t)}{dt}-4.21u(t)\cdot\left(2.1-0.15u(t)\right)=2.1t\cdot\cos(t)$ & [$1.23$, $17.02$] & $0.1$ & $10$\\
    ID29 & $\frac{du(t)}{dt}-0.1137u(t)=-9.7\sin(1.32t)$ & [$3.52$, $21.87$] & $12.3$ & $9$\\
    ID27 & $\frac{du(t)}{dt}-0.0837u(t)-\log\left(7.293u(t)\right)=0$ & [$1.8$, $8.27$] & $0.3$ & $9$ \\
    \hline
    \end{tabular}
    \label{tab:app1:add}
\end{table*}

\section{Details on the benchmark of linear and nonlinear first- and second-order ODEs}
\label{app:Bench2}

This section offers some additional details on the dataset generation and experiments of the second benchmark on implicit linear and nonlinear first- or second-order ODEs of \cref{ch3:subsec:bench2}.

The underlying grammar for the generation of the dataset is the probabilistic CFG in \cref{tab:app2:pcfg_bench2}. More complex rules are assigned lower probabilities to reduce the complexity of the generated expressions. The sum of all probabilities of a right-hand side for a specific left-hand side must sum up to 1.

\begin{table*}[h!]
    \caption{Probabilistic context-free grammar for generation of the dataset of the second benchmark.}
    \label{tab:app2:pcfg_bench2}
    \lstset{basicstyle=\small}
    \centering
    \begin{lstlisting}[breaklines]
        start -> linear_simple '*' Diff '*' Diff '*' Diff '-' '(' linear_simple ')' [0.02] | linear_simple '*' Diff '*' Diff '-' '(' linear_simple ')' [0.04] | expr '+' expr '-' '(' force ')' [0.34] | expr '+' expr '+' expr '-' '(' force ')' [0.4] | expr '+' expr '+' expr '+' expr '-' '(' force ')' [0.2]
        force -> '0' [0.4] | linear_simple [0.6]
        expr -> linear_simple '*' Diff [0.8] | linear_simple '*' Diff '*' Diff [0.15] | linear_simple '*' Diff '*' Diff '*' Diff [0.05]
        Diff -> 'diff' '(' 'diff' '(' 'u' ',' 't' ')' ',' 't' ')' [0.25] | 'diff' '(' 'u' ',' 't' ')' [0.25] | 'u' [0.5]
        linear_simple -> lin_all [0.8] | lin_all mop func [0.15] | lin_all mop func mop func [0.05]
        lin_all -> term [0.5] | term mop TV [0.5]
        TV -> 't' [0.75] | 't' '^' '2' [0.2] | 't' '^' '3' [0.05]
        func -> 'sin' '(' term '*' TV ')' [0.4] | 'cos' '(' term '*' TV ')' [0.4] | 'exp' '(' term '*' TV ')' [0.1] | 'log' '(' term '*' TV ')' [0.1]
        mop -> '*' [0.7] | '/' [0.3]
        term ->  'C' [1.0]
        Nothing -> None [1.0]
    \end{lstlisting}
\end{table*}

A more simplified context-free grammar is used for the GVAE in \cref{tab:app2:cfg_bench2}, which can also parse all generated expressions by the previous grammar.

\begin{table*}[h!]
    \caption{Context-free grammar underlying the GVAE for the second benchmark.}
    \label{tab:app2:cfg_bench2}
    \lstset{basicstyle=\small}
    \centering
    \begin{lstlisting}[breaklines]
        start -> expr '+' start | expr '-' force
        force -> '0' | linear_simple
        expr -> linear_simple '*' Diff | expr '*' Diff
        Diff -> 'diff' '(' Diff ',' 't' ')' | 'u'
        linear_simple -> lin_all | lin_all mop func | lin_all mop func mop func
        lin_all -> term | term mop TV
        TV -> 't' | 't' '^' '2' | 't' '^' '3'
        func -> 'sin' '(' term '*' TV ')' | 'cos' '(' term '*' TV ')' | 'exp' '(' term '*' TV ')' | 'log' '(' term '*' TV ')'
        mop -> '*' | '/'
        term ->  'C'
        Nothing -> None
    \end{lstlisting}
\end{table*}

\begin{figure*}[h!]
    \centering
    \includegraphics[width=\textwidth]{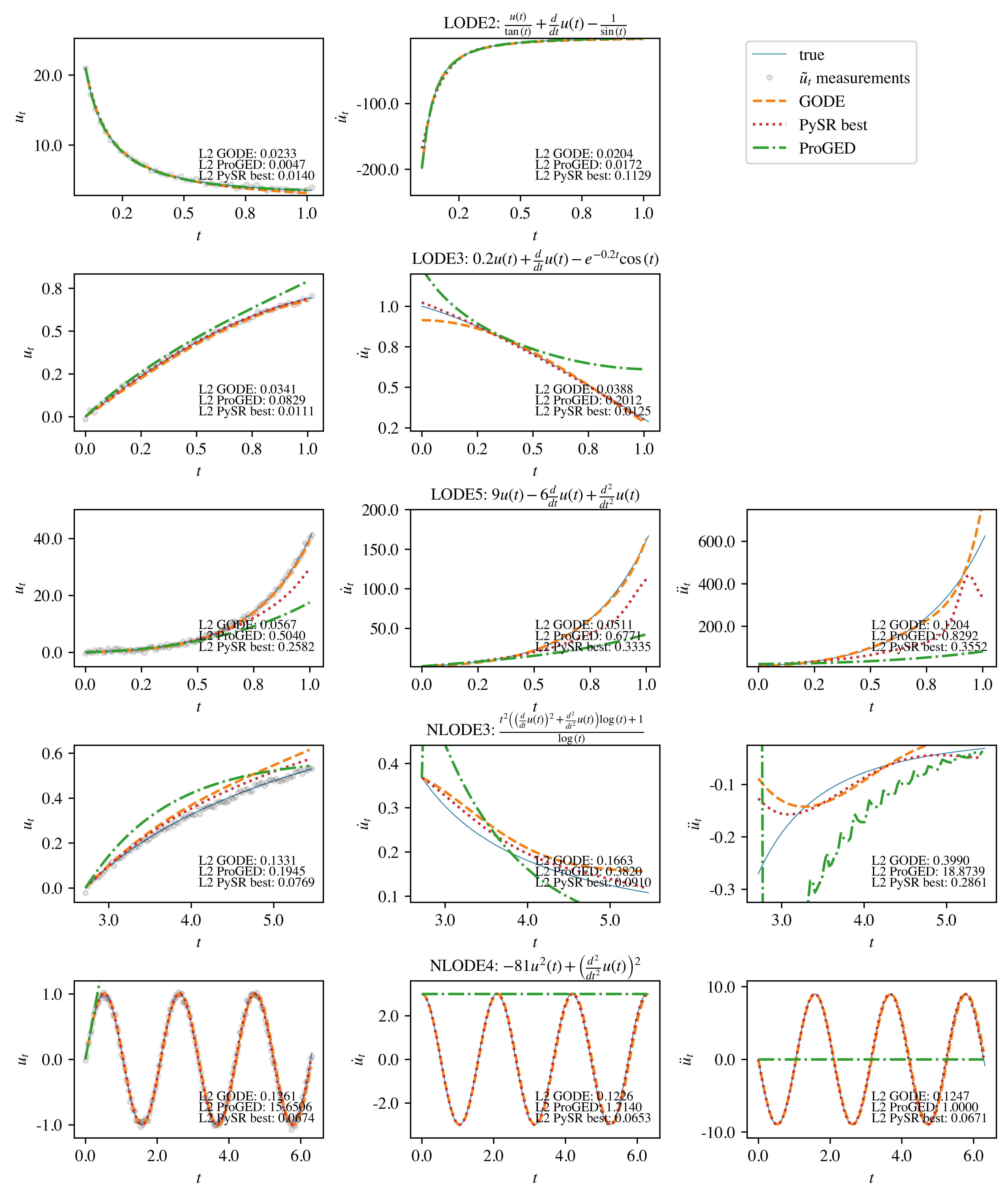}
    \caption{Comparisons between the system trajectories of the remaining predicted ODEs by GODE and the ground truth.}
    \label{fig:app2:bench2}
\end{figure*}

\cref{tab:app2:bench2} lists the ten tested examples with the time domain of $t$, the initial values, and sampling frequency. All of these examples have analytical solutions, which particularly for the nonlinear implicit ODEs are used to sample the solution trajectories. The comparison between the solution trajectories of the predicted and ground truth ODEs of the remaining examples are presented in \cref{fig:app2:bench2}.

\begin{table*}[h!]
    \renewcommand{\arraystretch}{1.75}
    \centering
    \caption{Test examples for LODEs and NLODEs. LODE1-4 and NLODE1-3 are from Tsoulos and Lagaris \cite{tsoulos_solving_2006}.}
    \begin{tabular}{|l|l|l|l|l|}
    \hline
    Name & Ordinary differential equation & Time domain & Init. values & Samp. freq. \\
    \hline
    \hline
    LODE1 & $\frac{du(t)}{dt}+\frac{u(t)}{2}-2=0$ & [$0.1$, $1.0$] & $20.1$ & $50$\\ 
    LODE2 & $\frac{du(t)}{dt}+u(t)\frac{\cos(t)}{\sin(t)}-\frac{1}{\sin(t)}=0$ & [$0.1$, $1.0$] & $\frac{2.1}{\sin(0.1)}$ & $50$\\ 
    LODE3 & $\frac{du(t)}{dt}+\frac{u(t)}{5}-\exp\left(-\frac{t}{5}\right)\cdot\cos(t)=0$ & [$0.0$, $1.0$] & $0.0$ & $50$\\ 
    LODE4 & $\frac{d^2u(t)}{dt^2}+64u(t)=0$ & [$0.0$, $1.0$] & $0.0$ & $100$\\ 
    LODE5 & $\frac{d^2u(t)}{dt^2}-6\frac{du(t)}{dt}+9u(t)=0$ & [$0.0$, $1.0$] & $0.0, 2.0$ & $100$\\ 
    NLODE1 & $\frac{du(t)}{dt}-\frac{1}{2u(t)}=0$ & [$1.0$, $4.0$] & $1.0$ & $30$\\ 
    NLODE2 & $\frac{d^2u(t)}{dt^2}\cdot\frac{du(t)}{dt}+\frac{4}{t^3}=0$ & [$1.0$, $2.0$] & $0.0$ & $50$\\ 
    NLODE3 & $\frac{d^2u(t)}{dt^2}\cdot t^2 + \frac{du(t)}{dt}^2\cdot t^2+\frac{1}{\log(t)}=0$ & [$e$, $2e$] & $0,\frac{1}{e}$ & $35$\\ 
    NLODE4 & $\frac{d^2u(t)}{dt^2}^2 -9 u(t)^2=0$ & [$0$, $2\pi$] & $0,3$ & $30$\\ 
    NLODE5 & $\frac{d^2u(t)}{dt^2}\cdot\frac{du(t)}{dt}-2.1u(t)-9.84t^3=0$ & [$0$, $2.5$] & $0,0$ & $50$\\ 
    \hline
    \end{tabular}
    \label{tab:app2:bench2}
\end{table*}

\section{Details on the benchmark of nonlinear dynamics ODEs}
\label{app:Bench3}

This section presents some additional details on the dataset generation of the third benchmark on engineering examples from nonlinear dynamics in \cref{ch3:subsec:bench3}.

The underlying grammar for the generation of the dataset is the probabilistic CFG in \cref{tab:app3:pcfg_bench3} and the grammar of the GVAE is listed in \cref{tab:app3:cfg_bench3}. As the force term is induced, no force term needs to be assembled with the grammar. 

\begin{table*}[h!]
    \caption{Probabilistic context-free grammar for generation of the dataset of the third benchmark.}
    \label{tab:app3:pcfg_bench3}
    \lstset{basicstyle=\small}
    \centering
    \begin{lstlisting}[breaklines]
        start -> linear_simple '*' Diff '*' Diff '*' Diff [0.02] | linear_simple '*' Diff '*' Diff [0.04] | expr '+' expr [0.34] | expr '+' expr '+' expr [0.4] | expr '+' expr '+' expr '+' expr [0.2]
        expr -> linear_simple '*' Diff [0.8] | linear_simple '*' Diff '*' Diff [0.15] | linear_simple '*' Diff '*' Diff '*' Diff [0.05]
        Diff -> 'diff' '(' 'diff' '(' 'u' ',' 't' ')' ',' 't' ')' [0.25] | 'diff' '(' 'u' ',' 't' ')' [0.25] | 'u' [0.5]
        linear_simple -> lin_all [0.8] | lin_all mop func [0.15] | lin_all mop func mop func [0.05]
        lin_all -> term [0.5] | term mop TV [0.5]
        TV -> 't' [0.75] | 't' '^' '2' [0.2] | 't' '^' '3' [0.05]
        func -> 'sin' '(' term '*' TV ')' [0.4] | 'cos' '(' term '*' TV ')' [0.4] | 'exp' '(' term '*' TV ')' [0.1] | 'log' '(' term '*' TV ')' [0.1]
        mop -> '*' [0.7] | '/' [0.3]
        term ->  'C' [1.0]
        Nothing -> None [1.0]
    \end{lstlisting}
\end{table*}

\begin{table*}[h!]
    \caption{Context-free grammar underlying the GVAE of the third benchmark.}
    \label{tab:app3:cfg_bench3}
    \lstset{basicstyle=\small}
    \centering
    \begin{lstlisting}[breaklines]
        start -> expr '+' start | expr
        expr -> linear_simple '*' Diff | expr '*' Diff
        Diff -> 'diff' '(' Diff ',' 't' ')' | 'u'
        linear_simple -> lin_all | lin_all mop func | lin_all mop func mop func
        lin_all -> term | term mop TV
        TV -> 't' | 't' '^' '2' | 't' '^' '3'
        func -> 'sin' '(' term '*' TV ')' | 'cos' '(' term '*' TV ')' | 'exp' '(' term '*' TV ')' | 'log' '(' term '*' TV ')'
        mop -> '*' | '/'
        term ->  'C'
        Nothing -> None
    \end{lstlisting}
\end{table*}

\clearpage

\end{document}